\documentclass[preprint,12pt]{elsarticle}
\biboptions{sort&compress}
\usepackage{graphicx}
\usepackage{hhline}
\usepackage{algorithm}
\usepackage[pagewise]{lineno}
\usepackage{multicol}
\usepackage{xcolor}
\usepackage{url}
\usepackage{multirow}
\usepackage{array}
\usepackage{amsmath}
\usepackage{amssymb}
\usepackage{threeparttable,booktabs}
\usepackage{etoolbox}
\appto\TPTnoteSettings{\footnotesize}
\usepackage{rotating} 
\usepackage[colorlinks = false,hidelinks]{hyperref}
\usepackage[utf8]{inputenc}
\usepackage[english]{babel}
\usepackage[paper=portrait,pagesize]{typearea}
\usepackage{pdflscape}
\usepackage{caption}
\usepackage{subcaption}

\hypersetup{pdfstartview={XYZ null null 1.00}}

\journal{ }

\begin{document}

\begin{frontmatter}

\title{Machine Learning to Predict the Antimicrobial Activity of Cold Atmospheric Plasma-Activated Liquids}

\author[IKCBME,IKCFBE]{Mehmet Akif Özdemir\fnref{contrib}}
\ead{makif.ozdemir@ikcu.edu.tr}

\author[IKCBME,IKCFBE]{Gizem Dilara Özdemir\fnref{contrib}}
\ead{gizemdilara.ekimci@ikcu.edu.tr}

\author[IKCFBE]{Merve Gül}
\ead{merve97gul@gmail.com}

\author[IKCBME]{Onan Güren}
\ead{onan.guren@ikcu.edu.tr}

\author[IKCBME]{Utku Kürşat Ercan\corref{corresponding}}
\ead{utkuk.ercan@ikcu.edu.tr}

\cortext[corresponding]{Corresponding author.}

\address[IKCBME]{Department of Biomedical Engineering, Izmir Katip Celebi University, Cigli, 35620, Izmir, Türkiye} 

\address[IKCFBE]{Department of Biomedical Technologies, Graduate School of Natural and
Applied Sciences, Izmir Katip Celebi University, Cigli, 35620, Izmir, Türkiye} 

\fntext[contrib]{These authors contributed equally to this work.}

\begin{abstract} 

Plasma is defined as the fourth state of matter and non-thermal plasma can be produced at atmospheric pressure under a high electrical field. The strong and broad-spectrum antimicrobial effect of plasma-activated liquids (PALs) is now well known. The antimicrobial effects of PALs depend on many different variables, which complicates the comparison of different studies and determines the most dominant parameters of the antimicrobial effect. The proven applicability of machine learning (ML) in the medical field is encouraging for its application in the field of plasma medicine as well. Thus, ML applications on PALs could present a new perspective to better understand the influences of various parameters on their antimicrobial effects. In this paper, comparative supervised ML models are presented by using previously obtained data to qualitatively predict the in vitro antimicrobial activity of PALs. A literature search was performed and data is collected from 33 relevant articles. After the required normalization, feature encoding, and resampling steps, two supervised ML methods, namely classification, and regression are applied to data to obtain microbial inactivation (MI) predictions. For classification, MI is labeled in four categories and for regression, MI is used as a continuous variable. 16 different classifiers and 14 different regressors are implemented to predict MI value. Two different robust cross-validation strategies are conducted for classification and regression models to evaluate the proposed method; repeated stratified k-fold cross-validation and k-fold cross-validation, respectively. We also investigate the effect of different features on models. The results demonstrated that the hyperparameter-optimized Random Forest Classifier (oRFC) and Random Forest Regressor (oRFR) provided better results than other models for the classification and regression, respectively. Finally, the best test accuracy of 82.68\% for oRFC and R\textsuperscript{2} of 0.75 for the oRFR are obtained. ML techniques could contribute to a better understanding of plasma parameters that have a dominant role in the desired antimicrobial effect. Furthermore, such findings may contribute to the definition of a plasma dose in the future.

\end{abstract}


\end{frontmatter}


\section{Introduction} \label{Introduction}

The term plasma was first introduced by Irving Langmuir in 1928 and refers to the fourth state of matter, which can be generated under an electric field and is a partially ionized gas composed of photons, free electrons, ions, free radicals, reactive oxygen species (ROS), and reactive nitrogen species (RNS)  \cite{moreau2008non}. Plasma can be produced at atmospheric pressure or under a vacuum, and can be classified into two categories based on thermal equilibrium in between the electrons and heavy particles; thermal or hot plasma and non-thermal or cold plasma. Cold atmospheric plasma (CAP) could be generated at atmospheric pressure and room temperature under an externally applied electrical field. CAP applications can be summarized in three main categories; surface modifications, therapeutic applications, and biological decontamination \cite{metelmann2018comprehensive}. Antimicrobial activity, blood clotting, tooth whitening, wound healing, and anticancer efficacy are the main biomedical applications of CAP that are reported in the literature \cite{oztan2022irrigation}. Due to its strong antimicrobial activity against a wide spectrum of microorganisms, including antibiotic-resistant organisms, CAP is an emerging technology that is undergoing intensive research. One of the treatment methods of CAP, direct CAP treatment has been used in sterilization and disinfection applications with the goal of microbial inactivation (MI) in a number of studies. Beside direct CAP treatment, liquids treated with plasma have been shown in the literature to have similar effects to direct CAP treatment by undergoing chemical modifications \cite{kim2021applications}. Reactive plasma-generated species such as ROS, RNS, free radicals, and electrons are transferred to the liquid by CAP treatment. Those reactive species may also lead to formation of new species via interaction with the treated liquid \cite{kaushik2019biological}. The liquid that is activated by CAP treatment is called plasma-activated liquids (PALs).

The plasma-generated reactive species were linked to the strong and broad-spectrum antimicrobial activity of PALs \cite{tsoukou2018understanding}. Xiang et al. conducted a comprehensive study on the antimicrobial efficacy of plasma-activated water (PAW) against a range of microbial strains \cite{xiang2020review}. Different liquids were activated via CAP treatment in another study \cite{ercan2013nonequilibrium}, including deionized water (DIW), N-acetyl-cysteine (NAC), and phosphate buffer saline (PBS), and these PALs exhibited broad-spectrum antimicrobial activity. Furthermore, this research reveals that PALs have a long-lasting antimicrobial effect. Schmidt et al. investigated the antimicrobial activities, pH value, and conductivity of several solutions (tap water, PBS, and physiological saline) that are activated with CAP \cite{schmidt2019plasma}. Studies in the field of MI by PALs are expanding and becoming more prevalent as a consequence of the advantages they provide. However, due to the differences between various plasma generation systems including electrical plasma parameters, electrode geometry, and type, liquid type, treatment volume, etc comparison of antimicrobial efficacies of PALs from different research laboratories can be difficult \cite{cheng2019colorimetric}. Furthermore, no well-established method for measuring, comparing, and predicting the antimicrobial effectiveness of PALs generated by different plasma systems has yet to be developed. Moreover, the inhibition activity of the produced PALs may vary depending on the liquids and microorganism strains. Thus, a method that would help to predict the antimicrobial strength and efficacy of a PAL is needed to achieve the desired antimicrobial activity by PALs.

Artificial intelligence (AI), a branch of computer science that is utilized in a wide range of disciplines and is receiving a lot of attention these days, is frequently employed in solving complicated problems, making decisions, and recognizing patterns \cite{mirzaei2021machine}. Machine learning (ML), one of AI's sub-branches, is the ability of constructed algorithms to learn from incoming data. ML is a data analysis technology that automates model construction bypassing in-depth comprehension and connecting input data to the final outcome. ML makes predictions by extracting inferences from data using mathematical and statistical operations. ML algorithms are used to identify models in data and to create a predictive data model with those models \cite{rajkomar2019machine}. In scenarios where classical statistical methods are insufficient, ML provides an excellent option for analyzing a very large amount of data due to its adaptability. AI is widely used in many fields nowadays since these tools are simple and affordable \cite{furxhi2019machine}. Furthermore, investigations based on this foundation offer light on their approach, which is based on data inputs rather than physical test materials. They can also anticipate the impacts of materials whose effects have yet to be discovered using constructed models, so contributing to prediction approaches without the need for complex optimization processes \cite{mirzaei2021machine}. Antibiotic development \cite{serafim2020application} and antimicrobial resistance prediction for specific bacteria \cite{liu2020evaluation} have both benefited from ML approaches. In the medical and biomedical disciplines, ML models are commonly used. The proven applicability of ML algorithms in the medical field bodes well for their use in other fields, such as the prediction of antimicrobial activity with different agents. There are only a few studies that apply ML techniques to estimate antimicrobial activity. Shaban et al. sought to estimate the antibiofilm activity of antibiotics using three optimized ML models based on logistic regression, decision tree, and random forest algorithms, using data manually obtained from the available literature for high-accuracy prediction of in vitro antibiofilm activity where they achieved 67\%±6.1 prediction accuracy for the logistic regression model, 73\%±5.8 for the decision tree model, and 74\%±5 for the random forest model \cite{shaban2022prediction}. Apart from that, there have been other studies on ML prediction involving nanoparticles and antimicrobial peptides. These studies collect data from the literature or databases and analyze them using ML algorithms to predict toxicity or antimicrobial activity \cite{furxhi2020practices,furxhi2020predicting}. 

In the field of plasma medicine, determining the plasma treatment parameters, attaining the required antimicrobial effect, and optimizing plasma treatment to a specific standardization is a critical challenge. To solve this problem, we propose an ML method to predict the antimicrobial activity of PALs against a wide spectrum of microorganisms. This tool uses CAP parameters, experimental setup parameters, and microorganism characteristics as inputs to predict antimicrobial efficacy. We compiled in vitro experimental data from research in the literature and structured them into a comprehensive dataset. By lowering the number of trial and error in laboratory studies, the current approach allows for the screening of PALs and the prediction of their ability to inactivate microorganisms, saving time and cost. To the best of our knowledge, no study has used ML algorithms to predict the antimicrobial efficacy of PALs. AI algorithms intended for use in the field of plasma medicine are expected to be pioneering studies in this field, paving the way for other researchers to use AI techniques in plasma medicine.

\section{Methods} 
\label{Methods}

The visual framework of the proposed methodological structure is presented in Figure \ref{Fig:framework}. A detailed description of each stage is given in the following.

\subsection{Data Collection}

A literature search was performed between December 2021 and March 2022 in the Scopus, ScienceDirect, PubMed, and Web of Science databases to find relevant articles on the antimicrobial activity of PALs which publication dates between 2008 and 2022. The keywords used in the literature search were determined as ‘‘antimicrobial, antibacterial, inactivation, cold atmospheric plasma, non-thermal plasma, cold plasma, plasma activated water, plasma activated liquid, plasma treated water, plasma treated liquid’’. A detailed expression of literature search is presented in Table S1 in Supplementary Data. 
In addition using keywords to search databases, articles were acquired by conducting a manual literature search as well. A total of 307 relevant articles were obtained, and the titles and abstracts of each article were examined to determine the most appropriate publications.

The primary inclusion criteria for the studies are to determine the MI value as a log reduction by exposing PAL to microorganisms. The characteristics of the microorganisms and cold plasma specifications utilized in the experiments should also be indicated in the article. The article removal process was carried out and a total of 33 papers were chosen. Afterward, the feature extraction stage was performed. Features such as CAP specifications, PAL characteristics, in vitro characteristics, and CAP treatment characteristics were manually extracted from the 33 articles \cite{ercan2013nonequilibrium, kojtari2013chemistry, hong2021antimicrobial, smet2019inactivation, shen2016bactericidal, zhao2020inactivation, li2017vitro, liu2021vitro, oehmigen2011estimation, kamgang2008impact, kamgang2009microbial, dezest2017oxidative, tsoukou2018understanding, simon2022influence, zhou2018cold, chiappim2021antimicrobial, schmidt2019plasma, hansch2015analysis, rathore2021investigation, joshi2018characterization, ma2020key, naitali2010combined, qi2018inactivation, royintarat2019mechanism, suganuma2018air, tian2015assessment, traylor2011long, wu2017reactive, xiang2019synergistic, Ye2013EfficiencyAM, zhang2013study, zhang2016sterilization, joshi2015microarray}. Any article which has missing value about the specified features was eliminated. As a result, a total of 762 observations were collected without any missing predictors value.

\subsection{Predictors and Outcome}
 
In the field of plasma medicine and antimicrobial studies with PALs plasma treatment type, gas type, discharge gap, plasma activated liquid type, treatment volume, treatment time, microbial strain, initial microbial load, PAL/microorganism suspension volume ratio, contact time, post-storage time, and incubation temperature are the most widely used parameters which were also used as predictors in the feature selection. Logarithmic values for MI were utilized for supervised ML models to relate these predictors to an outcome.

Before data preprocessing, features that were manually extracted from the relevant articles were modified. While discharge gap, treatment volume, treatment time, initial microbial load, PAL/microorganism suspension volume ratio, exposure time, post-storage, and the incubation temperature features of the PALs were numeric data, plasma type, gas type, liquid type, and microbial strain were nominal data among the features. The logarithm of the initial microbial load (\textit{n}) was also taken because one of the inclusion criteria for an article was to define the MI in logarithmic value. The input and output variables, their categories, and types as well as the total number of unique values of each variable are shown in Table \ref{table:predictors}. Also, the scatter plot of numerical predictors and outcome is presented in Figure S1 in Supplementary Data. 

MI value was also labeled and categorized to a certain standard for classification models. The MI value was determined based on the logarithm of the \textit{n} of the bacterial solution because the \textit{n} could vary from study to study. A computation approach was used to label the results as four different categorical data, as shown in Table \ref{table:MI}. Following the categorization of results, there are 211 observations in the ‘‘Complete’’ class, 94 in the ‘‘Strong’’ class, 169 in the ‘‘Weak’’ class, and 288 in the ‘‘None’’ class. The identified categorical classes were used to express the MI value for the classification models. Regression models were also utilized to estimate the output as an exact value. Therefore, MI was used as a continuous variable in regression models.

\subsection{Data Preprocessing}

The collected dataset for the creation of ML models was subjected to normalization, encoding, and resampling data preprocessing steps before the training in order to avoid some issues such as variability in unit sizes, data imbalance effect, and data type inconsistency.

\subsubsection{Normalization}

The variability in units of measurement and magnitudes during model training may make optimization more challenging. In order to improve model performance, the normalization technique was used as a pre-processing step \cite{ksikazek2019novel}. Data normalization is the process of converting the obtained data to distinct ranges. During the normalization step, Z-Score, Min-Max scaler \cite{pandey2017comparative}, Max-Absolute scaler, and Robust scaler \cite{ahsan2021effect} were implemented. Different normalization strategies have been tested taking into account that normalization may have an impact on the model's success because the majority of the data is numerical. All normalization methods were implemented for both classification and regression models. The best results for classification and regression were determined empirically among different normalization methods. The best normalization technique was adopted for the continuation of the experimental studies. The initial numerical data distribution as well as the updated numerical data distributions that were produced through the utilizing of the four distinct normalization approaches outlined above depicts in Figure S2 in Supplementary Data. 

\subsubsection{Feature Encoding}

The variables used in ML need to be converted into machine-readable data in order to train the model. The most popular method for achieving this transition among the several known techniques is the one-hot encoding which divides the column into multiple columns to transform the nominal data into numeric data. The nominal values are replaced by 1s and 0s, whether they have the specified feature or not \cite{yang2019identifying}. Since the frequency of the variables was not taken into consideration, nominal values that were repeated numerous times were only counted once. Because some of the predictors in this study contain nominal data and they have no rank or order, these features transformed to numeric values with the one-hot encoding approach. Contrary, the determined categorical outcome, MI, was encoded for classification models as gradually due to it having a rank from ‘‘Complete’’ to ‘None’’ classes. The ‘‘Complete’’ class is superior to the other classes so it was encoded as ‘‘3’’, and the ‘‘Strong’’, ‘‘Weak’’, and ‘‘None’’ classes were encoded as ‘‘2’’, ‘‘1’’, and ‘‘0’’, respectively.

\subsubsection{Resampling}

Classes are frequently uneven in studies using real data, such as the prevalence of diseases and experimental outcomes, and class imbalance is an important problem in studies using supervised ML \cite{tanha2020boosting}. Classifiers and regressors have less accurate predictions as a result of the large class difference. Correct classification and regression may be ensured by balancing the proportion of majority classes with minority classes \cite{alghamdi2017predicting}. There are a number of ML techniques, and numerous research claim that balanced data can improve prediction performance. A type of oversampling technique called Synthetic Minority Oversampling Technique (SMOTE) was used for classification to overcome the imbalance issue in unbalanced and high-dimensional datasets. The SMOTE technique creates new minority class samples based on the original dataset that was generated at random from the closest neighbors joining the minority class data to expand the amount of data \cite{chawla2002smote}. The label ‘‘None’’ predominates the dataset with 288 observations, and there are four categories of MI that were assigned as outcomes. All other MI categories should therefore be equivalent to 288. 

Synthetic Minority Over-Sampling Technique for Regression (SMOTER) with Gaussian Noise (SMOGN) is another resampling method to eliminate the imbalance problem \cite{branco2017smogn}. The SMOGN technique was used to resample the imbalanced dataset for use in the training of the regression models. The SMOTER and the introduction of Gaussian Noise are two oversampling techniques combined in the SMOGN approach. Based on the distance to the k-nearest neighbors, SMOGN iterates through all unusual samples and selects between SMOTER's interpolation-based oversampling and Gaussian noise-based oversampling \cite{steininger2021density}. The data imbalance effect was eliminated by using the SMOTE and SMOGN techniques for classification and regression models, respectively to resample the dataset into the classes and change the relative frequency of the other labels.

\subsection{Data Splitting}

To evaluate the robustness of the trained ML model dataset should be split \cite{ozdemirEEgEmotion2021}. Hence, unknown test data which never used in the training step can be predicted fairly with the trained ML model. Splitting the dataset is the last step before developing the model and may be grouped as train-test split and validation splitting. Train-test splitting in ML is the process of dividing the data sample into two groups, a ‘‘training set’’ and a ‘‘test set’’, for the purposes of training and testing the model \cite{yang2019identifying}. In this study, the data were divided into two groups for classification and regression models at random: a training set including 80\% of the data and a test set containing the remaining 20\%. It's also crucial to validate the outputs of the ML models to assure their accuracy. In this study, two different robust validation techniques namely repeated stratified k-fold cross-validation and k-fold cross validation were utilezed for classification and regression models, respectively by considering the training sets.

\subsection{Model Development}

Studies that use a range of ML techniques to create a variety of models are documented in the literature. These methods can be split into two groups: supervised learning and unsupervised learning. Supervised learning is the process of developing algorithms that reliably identify data or forecast outcomes using labeled datasets \cite{kaur2019diagnosis}. Supervised learning has two subcategories: regression and classification. Both supervised ML techniques were used in this study to predict the categorical or continuous MI value. Models were developed using Python version 3.7.0 and \textit{scikit-learn} version 1.0.2. 

\subsubsection{Classification Models}

In ML, the term ‘‘classification’’ refers to a predictive modeling problem where a class label is predicted for a given set of inputs. The ML method links CAP parameters, PAL characteristics, and in vitro characteristics to the inhibition of bacteria and makes it possible to predict the MI efficacy of PALs. As candidates for our model, we tested a number of supervised classification algorithms to see which one could provide the most accurate prediction.

Logistic Regression (LR), Na\"ive Bayes Classifier (NBC), k-Nearest Neighbor (KNN), Decision Trees (DT), Random Forest (RF), Support Vector Machines (SVM), and Boosting algorithms are some of the supervised learning methods. Understanding how a group of independent variables influence the result of the dependent variable is made easier with the help of LR \cite{christodoulou2019systematic}. A modest amount of training data is needed for the Na\"ive Bayes classifier in order to estimate the necessary parameters. When compared to other classifiers, they are reasonably fast \cite{xu2018bayesian}. A lazy learning technique called KNN saves every instance corresponding to the training set in n-dimensional space \cite{zhang2007ml}. A decision tree has the benefit of being easy to comprehend and depict, and it also requires a very small amount of data preparation. The decision tree's drawback is that it can produce complex trees that might not categorize things well. Due to less over-fitting than decision trees, the random forest has the advantage of being more accurate \cite{charbuty2021classification}. The random forest classifiers' only drawback is that their implementation is fairly difficult \cite{hengl2018random}. SVM is memory-efficient and works incredibly well in high-dimensional domains since it only employs a portion of the training points in the decision function. The SVM's sole drawback is that the technique does not elicit probability estimates directly \cite{sorensen2018ensemble}. Boosting is an ensemble learning technique that turns weak learners into strong learners in order to improve the model's accuracy. It makes use of a variety of ML algorithms \cite{rahman2020performance}.

LR, Linear Discriminant Analysis (LDA), Gaussian Process Classifiers (GPC), KNN, Decision Tree Classifier (DTC), Extra Trees Classifier (ETC), Random Forest Classifier (RFC), Gaussian Na\"ive Bayes (GNB), Bernoulli Na\"ive Bayes (BNB), SVM, Bagging Classifier (BC), Extreme Gradient Boosting Classifier (XGBC), AdaBoost Classifier (ABC), Histogram-based Gradient Boosting Classifier (HGBC), and Gradient Boosting Classifier (GBC) algorithms were utilized as classifiers to predict categorical MI in the model development stage of the study.

\subsubsection{Regression Models}

Regression is another subfield of supervised ML. It aims to create a model of the relationship between a certain number of features and a continuous target variable \cite{shipe2019developing}. One of the most fundamental kinds of regression in ML is Linear Regression (LR). A predictor variable and a dependent variable that are linearly related to each other compensate for the LR model. When there is a strong correlation between the independent variables, Ridge Regression (RR) is commonly used. Regularization and feature selection are both carried out through Least Absolute Shrinkage and Selection Operator (LASSO) Regression. It limits the regression coefficient's maximum absolute value. With a small change, Polynomial Regression (PR) is identical to multiple linear regression. The value of the regression coefficients is determined via Bayesian Regression using the Bayes theorem. Instead of locating the least squares, this regression method determines the posterior distribution of the features \cite{fahrmeir2021regression}. Furthermore, such algorithms DT, RF, boosting, and KNN can be adapted as regression models.

LASSO Regression, RR, Extreme Gradient Boosting Regression (XGBR), LASSO-Least-Angle Regression (LLars), k-Neighbors Regression (KNR), AdaBoost Regression (ABR), Extra Trees Regression (ETR), Bagging Regression (BR), Elastic Net Regression (ENR), Linear Support Vector Regression (LSVR), Bayesian Ridge Regression (BaR), Multi-Layer Perceptron Regression (MLPR), Random Forest Regression (RFR), and Gradient Boosting Regression (GBR) were used as regressors to determine which model could deliver the most accurate performance in the prediction of the exact value of MI.

\subsection{Hyperparameter Tuning}

Hyperparameters are the variables that define the model architecture. The model's parameters, which are learned during training, cannot be manually set. Starting with random parameter values, a model modifies them throughout the training process. Hyperparameters, on the other hand, are variables that are chosen before the model is trained \cite{schratz2019hyperparameter}. Hyperparameter values have the potential to increase or decrease model accuracy. Hyperparameter tuning is the process of searching for the ideal model architecture and its optimum parameters. It is a crucial phase in the model training process that allows the model to test various combinations of hyperparameters and make predictions using the optimal hyperparameter values \cite{degirmenci2021ECG}. For hyperparameter tuning, a variety of techniques are utilized, including grid search, random search, and informed search \cite{andonie2019hyperparameter}. A genetic algorithm is a method for hyperparameter tuning that is based on informed search and the real-world idea of genetics. The procedure begins by building a few models, selecting the best one, building other models that are similar to the best ones, and adding some randomness till the target accuracy is achieved. Grid and random search are both used in informed search and genetic algorithms. The \textit{tpot} library \cite{Olson2016EvoBio} predicts the optimum hyperparameter values, and the evolutionary algorithm chooses the best model after learning from past iterations. However, it takes a lot of computational sources to compute \cite{ippolito2022hyperparameter}. In this study, the \textit{tpot} classifier and \textit{tpot} regressor was combined with a genetic algorithm as an informed search technique for classification and regression models, respectively. Thus, in addition to empirical determining the best accurate classifiers and regressors, not only the model selection has been strengthened, but also the model's parameters have been tuned with the genetic algorithm-based informed search used as additional validation.

\subsection{Model Evaluation}

The robustness of the model was assessed using a variety of performance evaluation criteria. Accuracy (ACC), recall (REC), precision (PRE), F1-Score (F1), area under the receiver operating characteristic (ROC) curve (AUC) \cite{ozdemir2021classification}, Jaccard index (JI), and elapsed time (ET) \cite{mcguinness2010comparative} were selected as the evaluation criteria for classification models. The confusion matrix and ROC curve were also provided for the best classification model. The formulas of the selected performance metrics are presented as follows;

\begin{align}
\centering
ACC &= \frac{TP+TN}{TP+TN+FP+FN}\label{eqnACC}\\
REC &=\frac{TP}{TP+FN}\label{eqnREC}\\
PRE &=\frac{TP}{TP+FP}\label{eqnPRE}\\
F1 &=\frac{2\cdot TP}{2\cdot TP+FP+FN}\label{eqnF1S}\\
JI &=\frac{TP}{TP+FP+FN}\label{eqnJI}
\end{align}

where \textit{TP}, \textit{TN}, \textit{FP}, and \textit{FN} indicate predicted classes as True Positive, True Negative, False Positive, and False Negative, respectively.

To evaluate the regression models' performance, robust statistical metrics such as coefficient of determination or R-squared (R\textsuperscript{2}), mean absolute error (MAE), mean squared error (MSE), root-mean-squared error (RMSE) \cite{mirzaei2021machine}, and ET \cite{tohka2021evaluation} were calculated. The formulas of the selected performance metrics are presented below;

\begin{align}
\centering
R\textsuperscript{2} &= 1-\frac{\sum(y_{i}-\hat{y})^2}{\sum(y_{i}-\overline{y})^2} \label{eqnR2}\\
MAE &= \frac{1}{N}\sum_{i=1}^{N}|y_{i}-\hat{y}| \label{eqnMAE}\\
RMSE &= \sqrt{MSE} = \sqrt{\frac{1}{N}\sum_{i=1}^{N}(y_{i}-\hat{y})^2} \label{eqnRMSE}
\end{align}

where \textit{i} is the data point, \textit{N} is the number of data points, and $\hat{y}$ and $\overline{y}$ indicate the predicted value of actual value (\textit{y}) and mean value of \textit{y}, respectively.

In order to assess the robustness of the models, these evaluation metrics were calculated during the training, validating, and testing phases. Moreover, feature importance was given for the best classification and regression models. Feature importance analysis refers to techniques that calculate a weighted score for all the predictors for the best model. The scores simply represent the ‘‘importance’’ of each feature \cite{altmann2010permutation}. A higher score indicates that the particular attribute has a greater impact on the prediction model for MI. Furthermore, In order to determine whether there is a statistical significance in the performance of the optimized classification and regression models, a one-way ANOVA test was performed on each trained model's validation accuracies and R\textsuperscript{2} scores \cite{ozdemirHand2022}.

\section{Results}
\label{Results}

\subsection{Experimental Setup}

Data collection was performed in order to predict the antimicrobial efficacy of PALs using supervised ML techniques. The literature review was performed for research articles published between 2008 and 2022. 762 observations satisfied the required criterion without any missing value. For the development, training, validation, and testing of the ML models, these observations from the collected data were utilized.

Table \ref{table:statistics} represents statistical data for numeric variables. The discharge gap has a mean of 13.71±17.23 mm. Moreover, the range of this variable is between 0 to 81 mm. Plasma treatment time has a mean of 1165±2771.48 sec and a range of 0 to 14400 sec. Treatment volume ranges from 0.25 to 500 mL, with a mean of 45.52±116.56 mL. Another variable, the initial microbial load has a mean of 6.62±0.91 log, with a range of 2 to 9 log. PAL/mo suspension volume ratio, is one of the most crucial variables, with a mean value of 109.25±246.75 fold, ranging from 1 to 1000 fold. Contact time ranges from 0 to 1440 min and has a mean of 39.05±131.02 min. Incubation temperature has the mean of 23.5±12.4\textdegree C while the ranging from -80 to 50\textdegree C, having mostly positive values. The range of post-storage time is 0 to 6120 min, with a mean of 64.47±345.56 min. Additionally, the detailed expression of nominal predictors with their unique values' labels and frequencies in the total observations are presented in Table S2 in Supplementary Data. 
The most prevalent type for plasma treatment, which is a nominal predictor among the included articles and considering 762 observations, was vDBD with a frequency of 40.6\%. The second one was the plasma jet with a frequency of 33.3\%. Among the gas types, the air has a frequency of 78.5\% and was the most used gas type among the articles. The second most used gas type was argon+oxygen with a frequency of 14.6\%. When plasma-activated liquids were examined, DIW was the most used among the 33 articles. DIW has a frequency of 51.7\%, while the second most used liquid, saline, has a frequency of 18.4\%. Two model organisms, \textit{Escherichia coli (E.coli)} and \textit{Staphylococcus aureus (S.aureus)}, which are widely practiced in experiments, were the two species with the highest frequency in our study when microbial strain is considered. The most frequently used microorganism in the included studies was the gram-negative model organism \textit{E.coli}, which has a frequency value of 41.5\%. The second most frequently used microorganism was the gram-positive model organism \textit{S.aureus}, which has 31.1\% frequency. All nominal variables were selected to build ML models even though they have diverse data distribution and some variable frequency rates are higher than others in order to observe the impact and contributions on the predictions. 

\begin{figure}
\begin{center}
\includegraphics[width=\textwidth,height=\textheight,keepaspectratio]{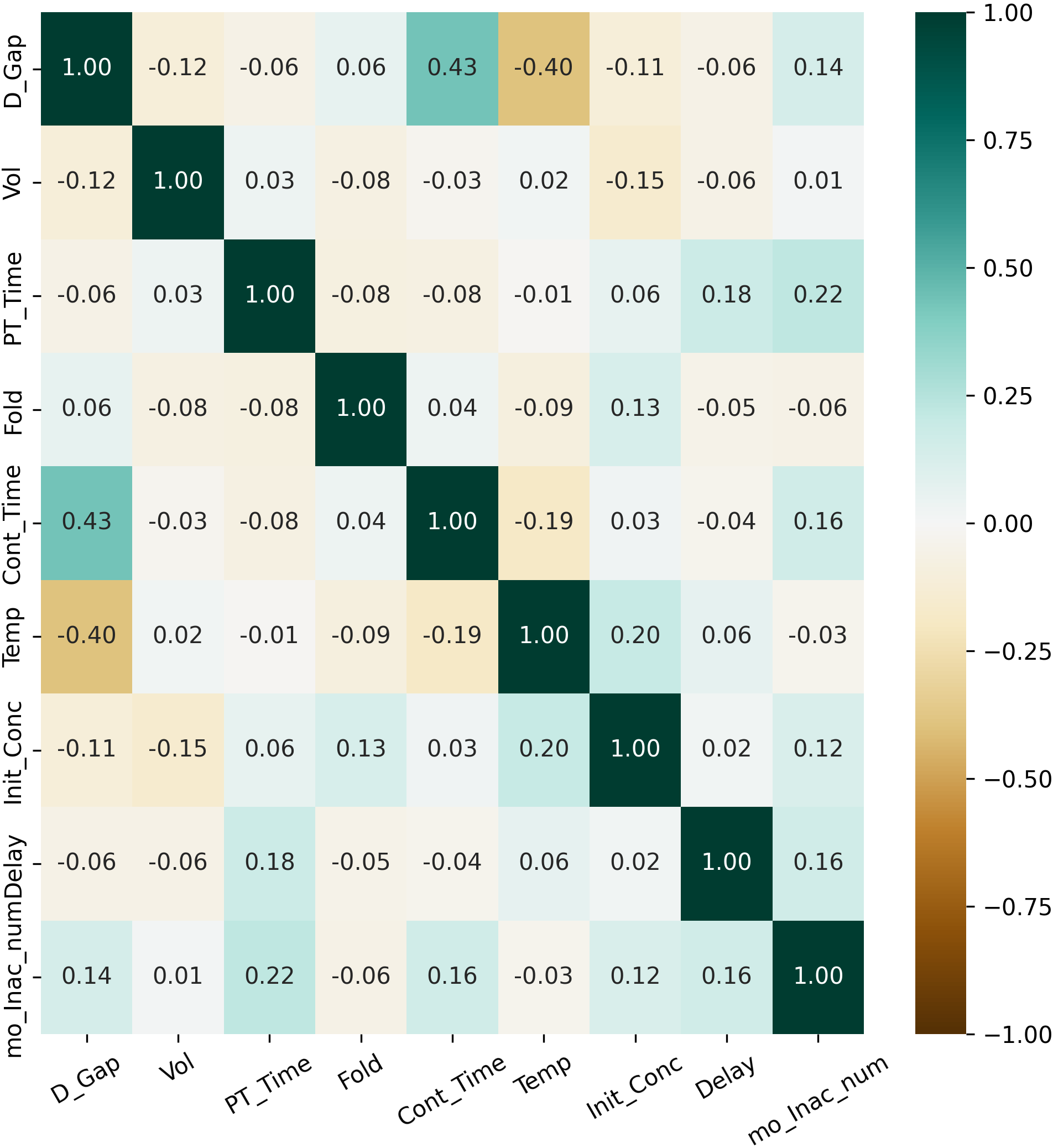}
\caption{Correlation matrix of predictors and outcome (D\_Gap: Discharge gap, Vol: Treatment volume, PT\_Time: Plasma treatment time, Fold: PAL/mo suspension volume ratio, Cont\_Time: Contact time, Temp: Incubation temperature, Init\_Conc: Initial microbial load, Delay: Post storage time, mo\_Inac\_num: Microbial inactivation numeric).}
      \label{Fig:heatmap}
\end{center}
\end{figure}

In order to better understand the relationships between the numeric predictors as well as outcome (MI), a correlation matrix was also created. A perceptible event is shown as color in two dimensions using the data visualization technique known as a correlation matrix. Based on hue or intensity, variation in color provides unmistakable visual cues as to how the event clusters or changes in space \cite{steiger1980tests}. Figure \ref{Fig:heatmap} showed the results of the correlation matrix. The correlation matrix revealed correlations between predictors and their correlations with output. Examining the correlation matrix reveals that there was generally no correlation between the predictors. Contact time, incubation temperature, and discharge gap might stand out as the most correlated features. The discharge gap and contact time have a positive correlation of 0.43 magnitude and the incubation temperature and discharge gap have a negative correlation of 0.40 magnitude. Therefore, no significant correlation between the predictors. As a result, all predictors might crucial contributions to the model development. Moreover, none of the predictors have a significant correlation with the output. Hence, similarly all predictors might contribute to the accurate prediction of the outcome. However, the expectation is that the most correlated predictors with the outcome should be most important for building the model. Therefore, all predictors were chosen for building the ML models, and contributions were examined. It should be also noted that the both collected numerical and nominal predictors are essential and key variables that a significant biological impact on MI.

Preprocessing steps were used to prepare the data for ML models by distributing the data evenly, converting nominal data to numeric data, and avoiding data imbalance. After normalization and resampling, a total of 1152 observations were obtained. The pre-processed data was randomly separated at a ratio of 80:20 ratio as a subset of the training and test set. 80\% of the data (922 observations) and 20\% of the data (230 observations) were used for training and testing, respectively. For classification and regression models, 10\% (92 observations) of the training data was used for validation in each fold. Repeated stratified 10-fold cross-validation was used for classification models as a validation strategy and ensured that arguments from the same essay are not distributed over the train, test, and development sets. The cross-validation was repeated 3 times which yields a total of 30 folds. For regression models, a 10-fold cross-validation strategy was used. Following the model development, the hyperparameter tuning step was utilized to enhance the model's performance and check once more which model had the highest accuracy in predicting MI. The \textit{tpot} classifier was employed for the optimization of the hyperparameters utilized in the models in order to improve their accuracy. For both the genetic algorithm-based \textit{tpot} classifier and regressor, the generation size was set to 100, the population size to 150, and the offspring size to 20. Both informed searches were conducted with a traditional cross-validation strategy to determine the best pipeline. An Intel\textsuperscript{\textregistered} Core\textsuperscript{TM} i9-10940X CPU and 64 GB RAM were used to develop all supervised ML models.

\subsection{Classification Results}

\begin{figure}
\begin{center}
\includegraphics[width=\textwidth,height=\textheight,keepaspectratio]{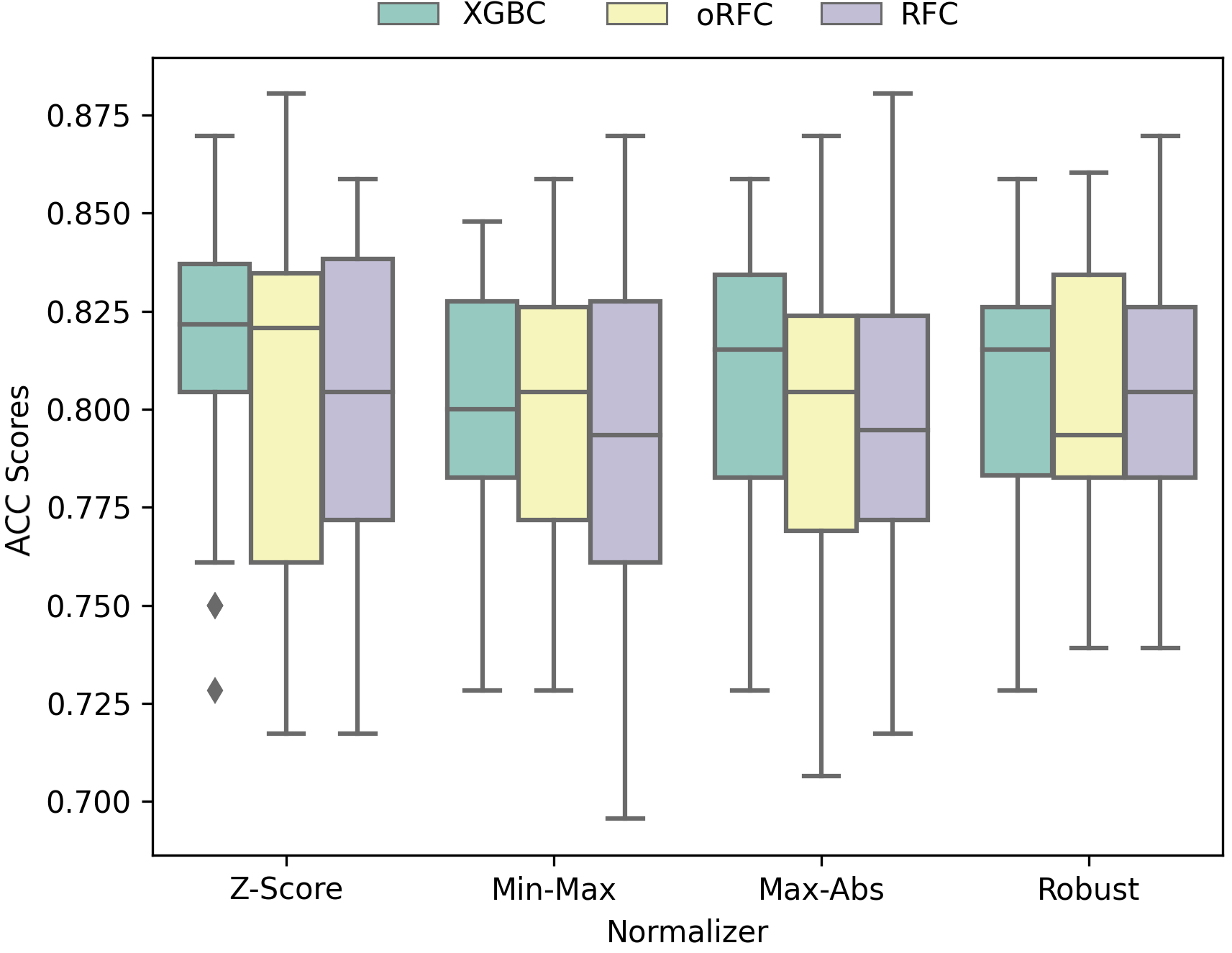}
\caption{A box-plot representation of accuracy values of different normalization methods with the top three classifiers by using a 3-repeated stratified-10-fold cross-validation strategy (The line in the boxes indicates the median value of ACCs).}
      \label{Fig:clf_normalization_results}
\end{center}
\end{figure}

To enhance the accuracy of the models, the numeric input variables underwent data normalization. Several methods, including Z-score, Min-Max scaler, Max-Absolute scaler, and Robust scaler were used as normalization techniques. The performance of three distinct classifiers and the various normalization techniques were compared. Figure \ref{Fig:clf_normalization_results} presents the accuracy scores obtained from the XGBC and RFC with unoptimized hyperparameters, and optimized RFC (oRFC) models by using a 3-repeated stratified-10-fold cross-validation strategy. oRFC is a hyperparameter-optimized version of conventional RFC by derived from the \textit{tpot} classifiers. The average accuracy values after applying the Z-score normalization technique were 80.38\%, 80.56\%, and 81.29\% for the XGBC, RFC, and oRFC models, respectively. The classifier with the highest accuracy value was oRFC, which accuracy value before normalization was 70.11\% and a huge increase to 81.29\% after normalization. Such a high accuracy change shows the effect of the magnitude difference of the numerical values on the performance of the model. Additionally, while oRFC yielded average accuracy of 81.29\% using Z-score normalization, 79.37\%, 79.87\%, and 80.35\% accuracies were obtained by utilizing Min-Max scaler, Max-Absolute scaler, and Robust scaler, respectively.  This accuracy difference enabled Z-Score selection among others and to be implemented in the next stages of the study.

\begin{figure}
\begin{center}
\includegraphics[width=\textwidth,height=\textheight,keepaspectratio]{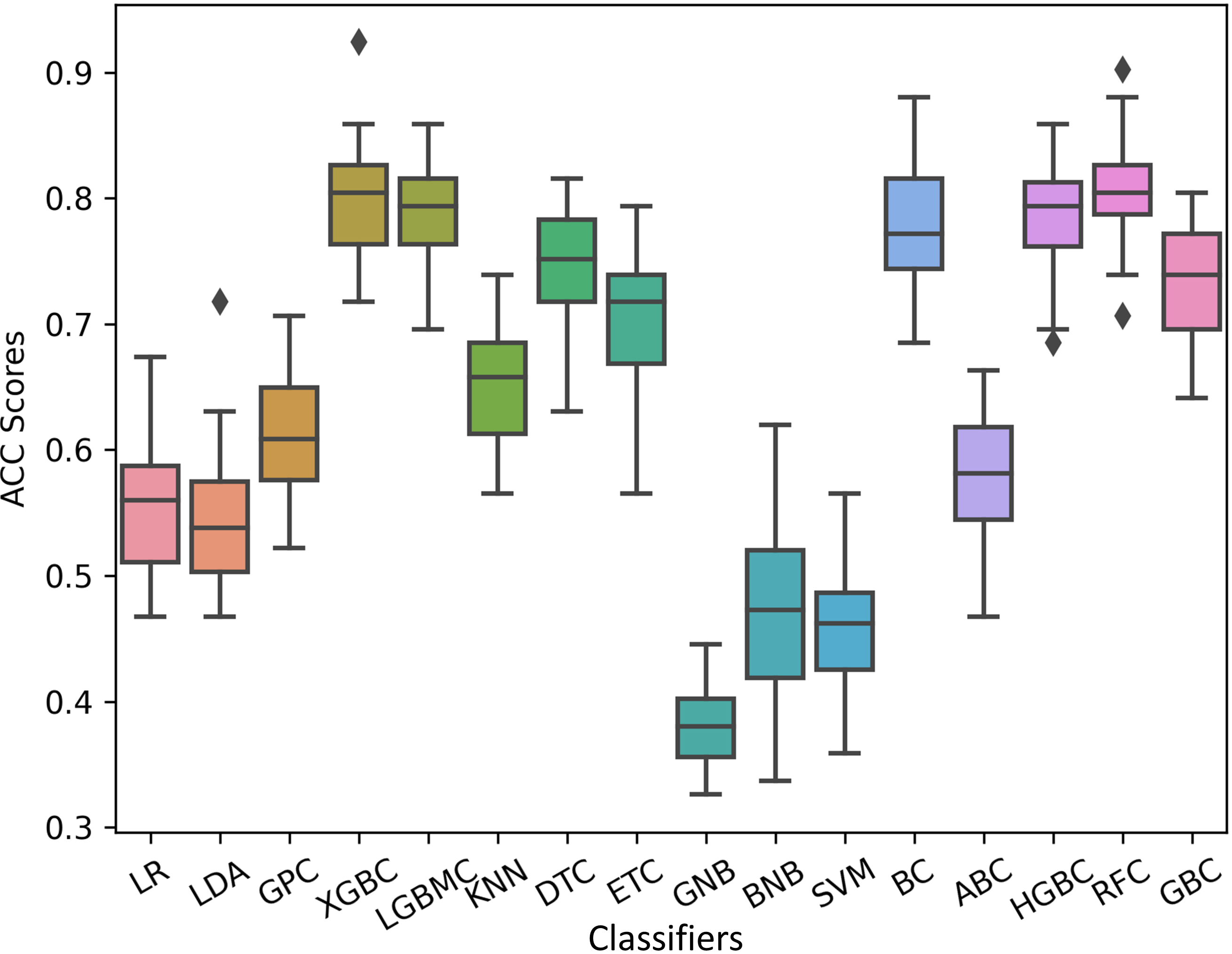}
\caption{A box-plot representation of obtained validation accuracy values of different classifiers by using a 3-repeated stratified-10-fold cross-validation strategy (The line in the boxes indicates the median value of ACCs).}
      \label{Fig:classifier_results}
\end{center}
\end{figure}

Various classifiers were utilized to choose the best accurate one among them. The different classifier results were presented in Figure \ref{Fig:classifier_results} as classifiers versus accuracy scores. RFC, XGBC, and LGBMC were the first three classifiers which led to average validation accuracy scores of 80.47\%, 80.23\%, and 78.63\%, respectively. The three classifiers with the lowest validation accuracy scores were GNB, SVM, and BNB, with average accuracy values of 38.17\%, 46.20\%, and 46.90\%, respectively. For this study, RFC is the most accurate model that may provide the most precise predictions. It should be noted that the accuracy values may vary due to the stochastic nature of ML models and the fact that the dataset randomly shuffles each time. While the accuracy value of the RFC model in the normalization comparison was 80.56\%, the accuracy value in comparing classifiers was 80.47\% with acceptable variance changing. Table \ref{table:classifiermetrics} summarizes the performance metrics of various classifiers for validation and training results. ACC and AUC were provided for training results and ACC, F1, REC, PRE, JI, AUC, and ET were provided for validation results. Please note that F1, REC, and PRE values were provided as macro scores. There was a 42.3\% accuracy difference between the most accurate model (RFC) and the least accurate model (GNB). To evaluate whether a model has statistical significance, a one-way ANOVA test was performed and the $p-value$ threshold was determined as .001. Statistical analyses demonstrated that while RFC and XGBC accuracy scores were statistically significant ($p<.001$) among other classifiers, no significance ($p>.999$) was found for other classifiers. Despite LGBMC and HGBC having the accuracy value of 78.63\% and 78.50\%, which is close to the RFC accuracy score, there was no statistical difference for both classifiers. For the RFC, the average training ACC and AUC were 99.00\% and 1.00, respectively. The validation F1-score and REC were 80.47\%, PRE was 81.03\%, JI was 0.68, and AUC was 0.94. In the second most accurate model, XGBC results, the ACC values are in line with the RFC. But, XGBC yielded better ET than RFC with 9.76 s. RFC was determined as the best classification model and used for the hyperparameter tuning and testing phase of the study. The statistical tests also supported the superiority of the RFC algorithm among others. 

After hyperparameter tuning, oRFC was obtained by using the predefined parameters to improve model accuracy. By optimizing the RFC model using the parameters obtained by the \textit{tpot} classifier, a new model was developed. The optimized parameters were determined such learning rate was 0.1, maximum depth was 10, maximum features was 0.3, minimum samples leaf was 7, minimum samples split was 10, estimators count was 100, and a subsample was 0.85 for the oRFC model. Prior to hyperparameter tuning, the accuracy value for RFC was 79.29\%. After hyperparameter tuning, the accuracy value for the oRFC model was enhanced to 81.29\%. Model accuracy was clearly impacted by normalization methods and hyperparameter tuning. The oRFC model was tested using the 20\% test data that was allocated during the data splitting stage after the model's performance had been established. In the test phase, the oRFC model was achieved 82.68\% ACC, 82.55\% F1 macro, 82.47\% REC, and 82.71\% PRE value. Furthermore, JI, AUC, Cohen's kappa, and Matthews correlation coefficients were calculated as 0.71, 0.88, 0.77, and 0.77, respectively. A visual representation of the tree structure of the oRFC model is presented in Figure S3 in Supplementary Data. 

\begin{figure}
\begin{center}
\includegraphics[width=\textwidth,keepaspectratio]{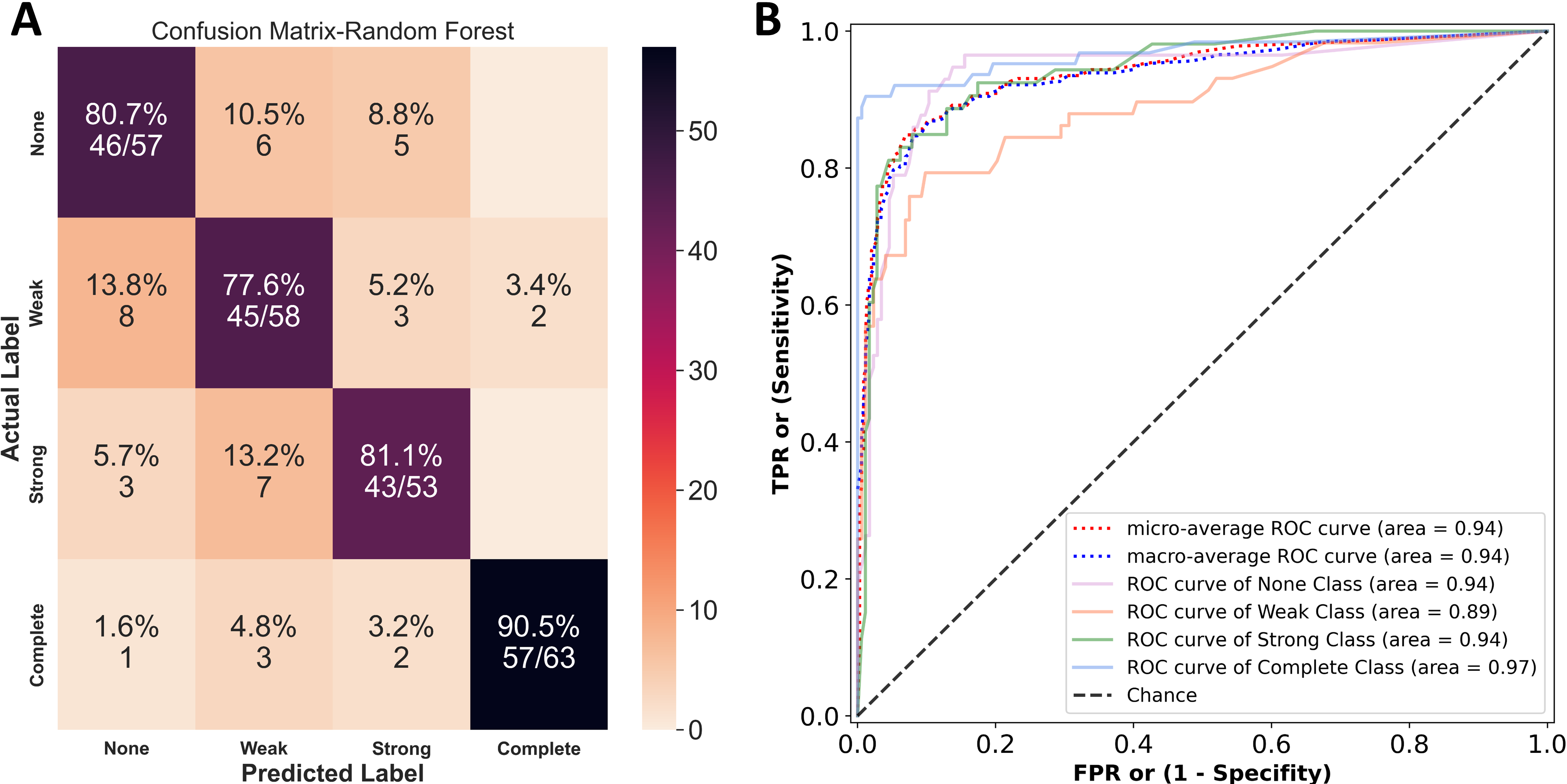}

\caption{Result graphics of the test phase (model predict) of the eventual oRFC model: (a) Confusion matrix and (b) ROC curve.}\label{Fig:confusion_ROC}
\end{center}
\end{figure}

In this study, a confusion matrix was employed to assess the effectiveness of the classification model. By comparing the actual values with those predicted by the trained model, the confusion matrix evaluates how the classification model works well\cite{haghighi2018pycm}. Figure \ref{Fig:confusion_ROC}a shows the confusion matrix created using the preprocessed dataset with the oRFC model. The distribution of the four indicators in the dataset—TP, FP, TN, and FN—was shown in the confusion matrix. The confusion matrix was generated in the testing phase with the use of the previously indicated four output labels: None, Weak, Strong, and Complete. The confusion matrix's ‘‘TP’’ and ‘‘TN’’  values were those instances where the expected value and the actual value match. The accuracies of the four output labels for the oRFC model were 80.7\% for ‘‘None’’, 77.6\% for ‘‘Weak’’, 81.1\% for ‘‘Strong’’, and 90.5\% for ‘‘Complete’’. The estimated values have high match percentages with the true values, which were represented by the darker boxes, and low match percentages with the mismatched values, which were represented by the lighter boxes, proving that the chosen model is a reliable model. The overall test accuracy was yielded as 82.68\%. Figure \ref{Fig:confusion_ROC}b shows the ROC curve that compares the rate of FP values to TP values. The gold standard of the classification has the largest AUC, and the sensitivity of the model rises as this value approaches 1. The area under the ‘‘Complete’’ class' curve with a value of 0.97 has the most sensitivity, whereas the area under the ‘‘Weak’’ class with a value of 0.89 has the lowest. Finally, the average test AUC score was computed as 0.94.

\begin{figure}
\begin{center}
\includegraphics[width=\textwidth,height=\textheight,keepaspectratio]{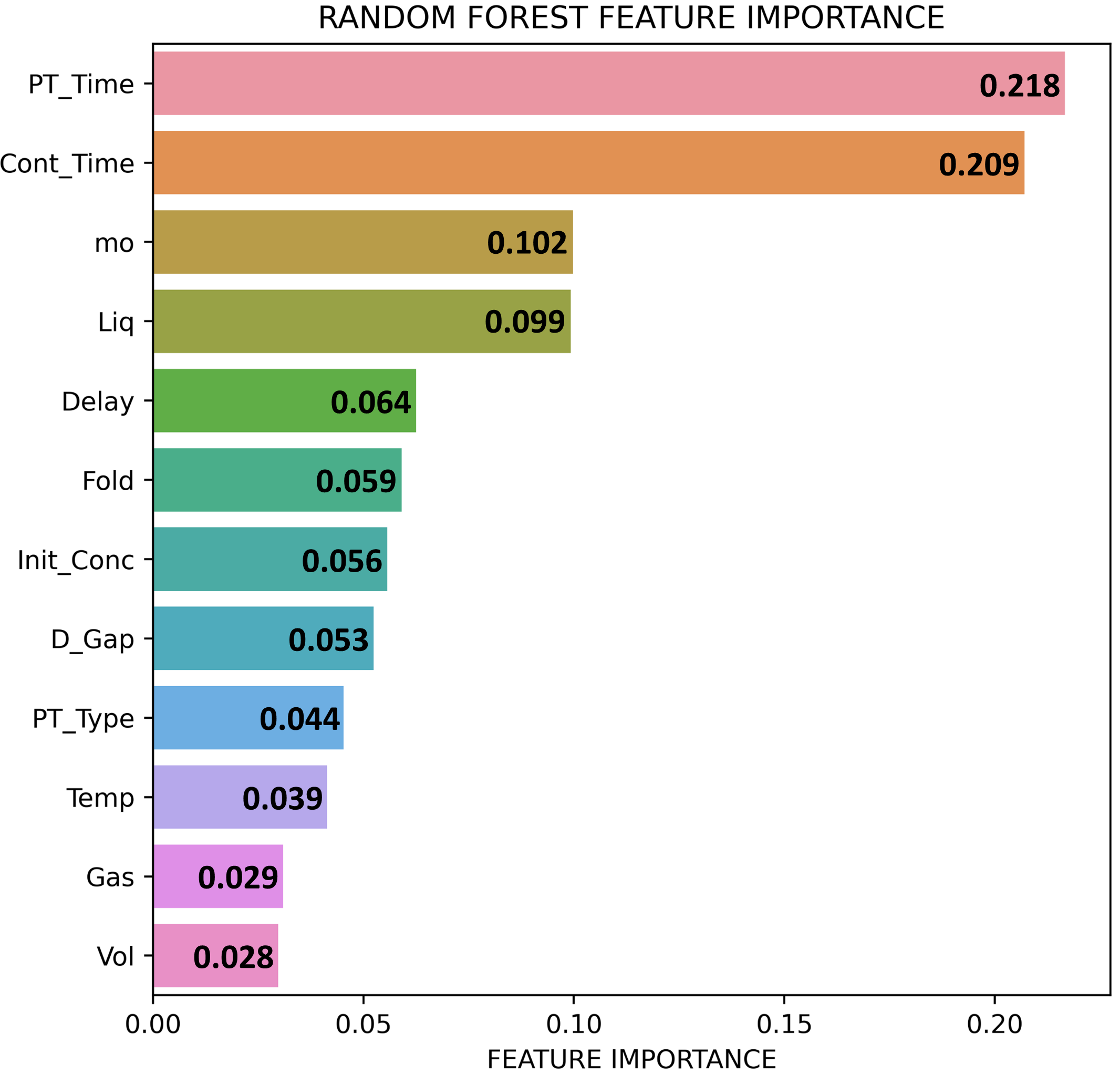}
\caption{Feature (predictor) importance results for eventual oRFC model. Feature importances were determined by considering the feature's weight wherein the final classification model function (PT\_Time: Plasma treatment time, Cont\_Time: Contact time, mo: Microbial strain, Liq: Plasma activated liquid, Delay: Post storage time, Fold: PAL/mo suspension volume ratio, Init\_Conc: Initial microbial load, D\_Gap: Discharge gap, PT\_Type: Plasma treatment type, Temp: Incubation temperature, Gas: Gas type, Vol: Treatment volume).}
      \label{Fig:clf_feature_importance}
\end{center}
\end{figure}

The final analysis for the classification model is presented in Figure \ref{Fig:clf_feature_importance} which represents the findings of the feature or predictor importance analysis. The weights of the predictors' coefficients, whichin the ML model function, were used to construct a measure of feature importance. Then the predictors were ordered starting with the highest coefficient. The highest weight represents the most contributed feature to the building of the best oRFC model. Hence, the greatest ones have a tremendous impact on the accurate prediction of MI. In comparison to other variables, plasma treatment time, contact time, microbial strain, liquid type, and post-storage time were identified as more significant features; it may be inferred that these attributes have the greatest impact on the output. Despite the effects of treatment volume, gas type, incubation temperature, plasma treatment type, and discharge gap were not as significant as the other features, the distributions of coefficients were not only weighted on certain ones. This indicates all predictors may have a significant impact on not only the building of the classification model but also on the accurate prediction of MI. These results were also in line with the correlation map results of the numerical data. 

\subsection{Regression Results}

\begin{figure}
\begin{center}
\includegraphics[width=\textwidth,height=\textheight,keepaspectratio]{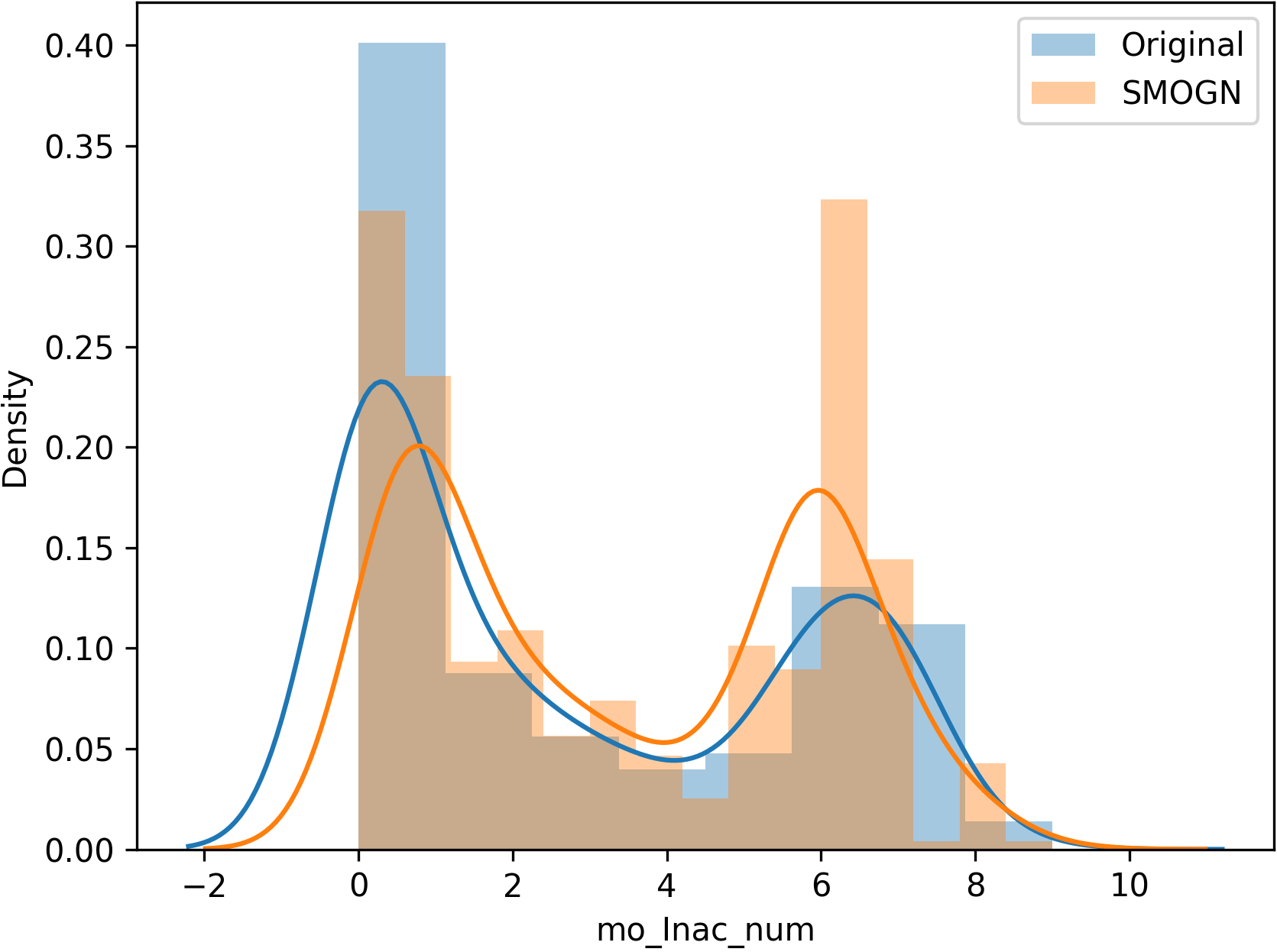}
\caption{The data distribution graph of before and after the SMOGN method (mo\_Inac\_num: Microbial inactivation numeric).}
      \label{Fig:smogn_balance}
\end{center}
\end{figure}

The dataset was resampled by using SMOGN as a data preprocessing step before the training of the regression models because the data distribution was not balanced. The SMOGN algorithm, which mixes oversampling with Gaussian noise \cite{steininger2021density}, is based on the SMOTER method. The density of the outcome of the dataset after oversampling with the SMOGN approach is shown in Figure \ref{Fig:smogn_balance} together with the initial data density. The majority of the original data distribution ranged between 0 and 2 for MI. Following the application of SMOGN, the data was distributed more uniformly and nearly in a Gaussian distribution. The data imbalance problem was reduced using the SMOGN technique, and then all data was resampled to produce a homogenous Gaussian data distribution. The SMOGN implementation has affected R\textsuperscript{2} scores. While the average R\textsuperscript{2} value was 0.68 before resampling, with the 0.04 increments, this value increased to 0.72 after the SMOGN application. Although the effect of the SMOGN on the validation R\textsuperscript{2} scores seems not too significant, the resampling method might make a clear impact on the predicting of new incoming data. Therefore, the importance of the resampling method is undeniable.

\begin{figure}
\begin{center}
\includegraphics[width=\textwidth,height=\textheight,keepaspectratio]{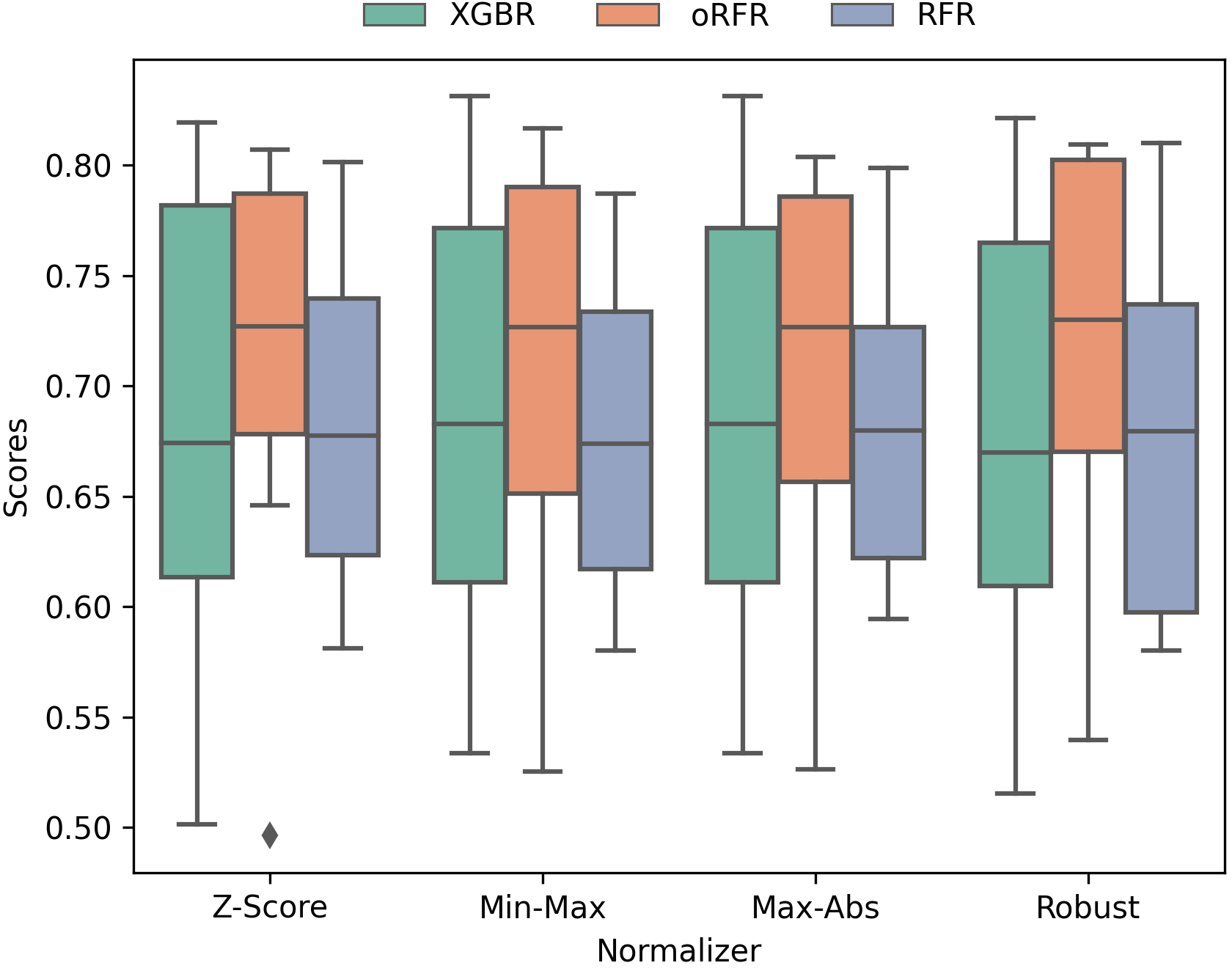}
\caption{A box-plot representation of R\textsuperscript{2} scores of different normalization methods with the top three regressors by using a 10-fold cross-validation strategy (The line in the boxes indicates the median value of R\textsuperscript{2}s).}
      \label{Fig:reg_normalizer_results}
\end{center}
\end{figure}

To enhance regression performance, similar to the classification problem, normalization methods like Z-Score, Min-Max, Max-Absolute, and Robust scaler were performed. The highest-performance top three regressors were chosen to compare the effect of the normalization techniques to the R\textsuperscript{2} scores. Figure \ref{Fig:reg_normalizer_results} presents the R\textsuperscript{2} scores obtained from the XGBR and RFR with unoptimized hyperparameters, and hyperparameter-optimized version of RFR (oRFR) models by using a traditional 10-fold cross-validation strategy. The oRFR which is a  similar version of the oRFC model was derived by using the \textit{tpot} regressors. The average R\textsuperscript{2} values after applying the Robust scaler normalization technique were 0.68, 0.68, and 0.72 for the XGBR, RFR, and oRFR models, respectively. The regressor with the highest average R\textsuperscript{2} score was oRFR, which R\textsuperscript{2} value before normalization was 0.64 and increased to 0.72 after normalization. While the average R\textsuperscript{2} value was achieved for the best regressor, oRFR, as 0.72, other normalization methods yielded the average R\textsuperscript{2} scores as 0.71 for all three normalization methods (Z-Score, Min-Max, and Max-Abs). The Robust scaler normalization method was selected to be implemented in the next stages of the study since it was the most effective method on the R\textsuperscript{2} scores.

\begin{figure}
\begin{center}
\includegraphics[width=\textwidth,height=\textheight,keepaspectratio]{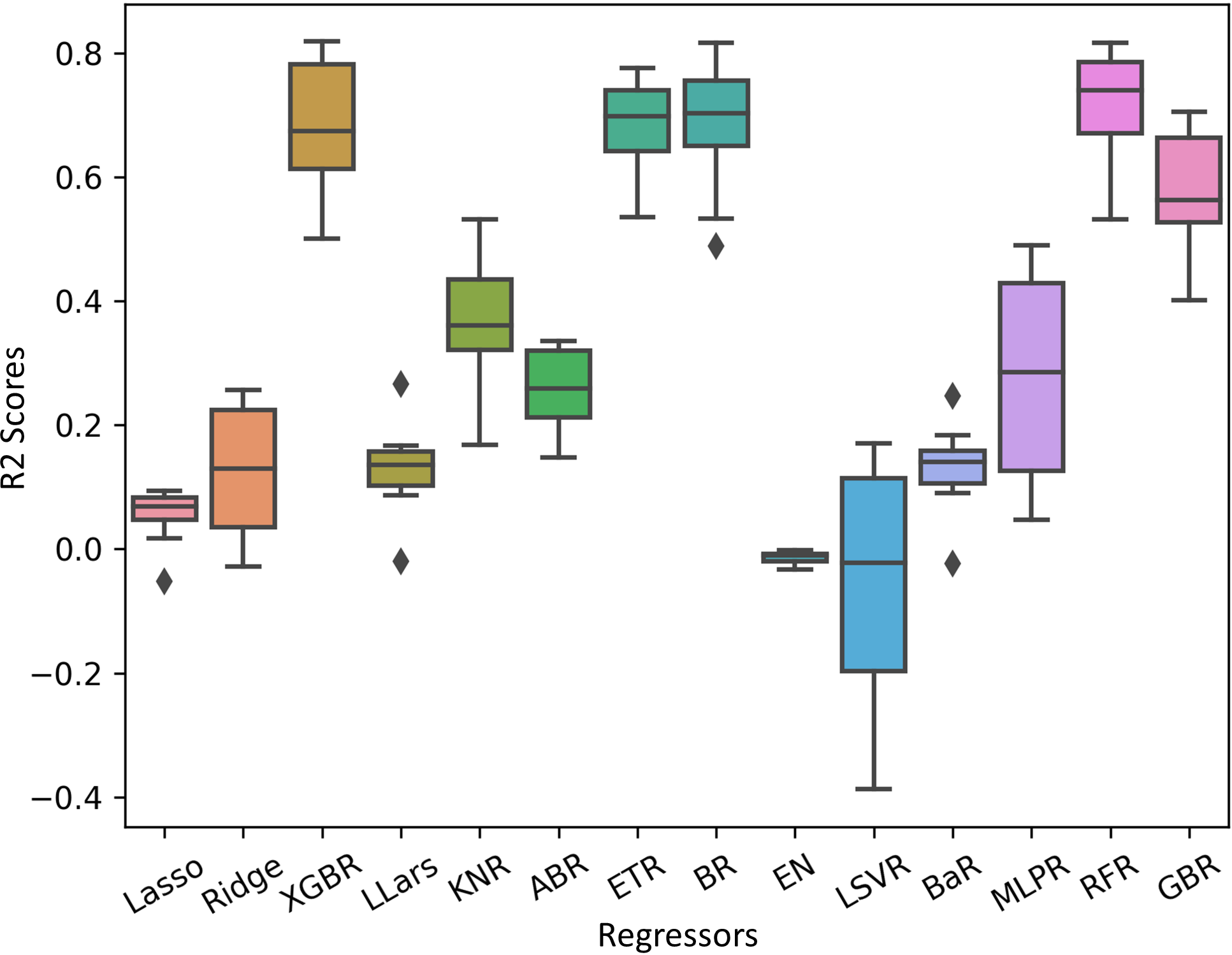}
\caption{A box-plot representation of obtained validation R\textsuperscript{2} scores of different regressors by using a 10-fold cross-validation strategy (The line in the boxes indicates the median value of R\textsuperscript{2}s).}
      \label{Fig:regressor_results}
\end{center}
\end{figure}

Following the data preprocessing steps, regression models were trained for the optimized dataset. The various regressor results were presented in Figure \ref{Fig:regressor_results} as regressor versus R\textsuperscript{2} scores. The three best R\textsuperscript{2} scores belong to RFR, BR, and ETR with 0.72, 0.69, and 0.68 scores, respectively. The three regressors with the lowest validation R\textsuperscript{2} scores were ENR, LASSO, and LSVR, with the average R\textsuperscript{2} values of 0.00, 0.05, and 0.06, respectively. Same as classification results, the RF-based regressor was the most accurate one. Due to the stochastic nature of the model, R\textsuperscript{2} values may vary between normalization and regression results similar to the classification scenarios. Table \ref{table:regressormetrics} summarizes the performance metrics of conventional 10-fold cross-validation strategy for various regressor. While R\textsuperscript{2}, MAE, and RMSE were provided for training results, R\textsuperscript{2}, MAE, MSE, RMSE, and ET were provided for validation results. According to the results, the RFR model has the lowest error and the greatest R\textsuperscript{2} score when compared to the other regressors with average R\textsuperscript{2}, MAE, MSE, and RMSE values of 0.72, 0.33, 0.28, and 0.53, respectively. There was a 0.72 R\textsuperscript{2} value difference between the most accurate model (RFR) and the least accurate model (ENR). One-way ANOVA was performed and results demonstrated that while RFR and BR model's R\textsuperscript{2} scores were statistically significant ($p<.001$), no significance ($p>.999$) was found for other regressors. Despite ETR and XGBR models both having the R\textsuperscript{2} value of 0.68, which is close to the RFR R\textsuperscript{2} score, there was no statistical difference for both regressors. For the RFR, the average training R\textsuperscript{2}, training MAE, and training RMSE were 0.95, 0.13, and 0.23, respectively. In the second most accurate model, BR has the R\textsuperscript{2} value of 0.69, which is 0.03 lower than RFR. MAE, MSE, and RMSE values were near to RFR. However, BR yielded better ET than RFR with 0.49 s. RFR was determined as the best regression model and used for the hyperparameter tuning and testing phase of the study. Similar to the classification results, the statistical tests also supported the superiority of the RF-based algorithm among others.

By employing the predefined parameters to increase model R\textsuperscript{2} score after hyperparameter tuning, the oRFR model was produced. The optimum parameters for the oRFR model were as follows: a subsample of 0.75, a learning rate of 0.01, maximum depth of 8, maximum features of 0.5, minimum samples leaf of 8, minimum samples split of 12, estimators count of 100. Before hyperparameter tuning, the RFR model's R\textsuperscript{2} score was 0.68. The R\textsuperscript{2} value for the oRFR model was increased to 0.72 after hyperparameter tuning. The oRFR model was tested using the 20\% test data that was allocated during the data splitting stage after the model's performance had been established. In the test phase, the oRFR model was achieved nearly 0.75 R\textsuperscript{2}, 0.32 MAE, 0.25 MSE, and 0.50 RMSE value. Furthermore, the maximum error and variance scores were calculated as 1.72 and 0.75, respectively.

\begin{figure}
\begin{center}
\includegraphics[width=\textwidth,height=\textheight,keepaspectratio]{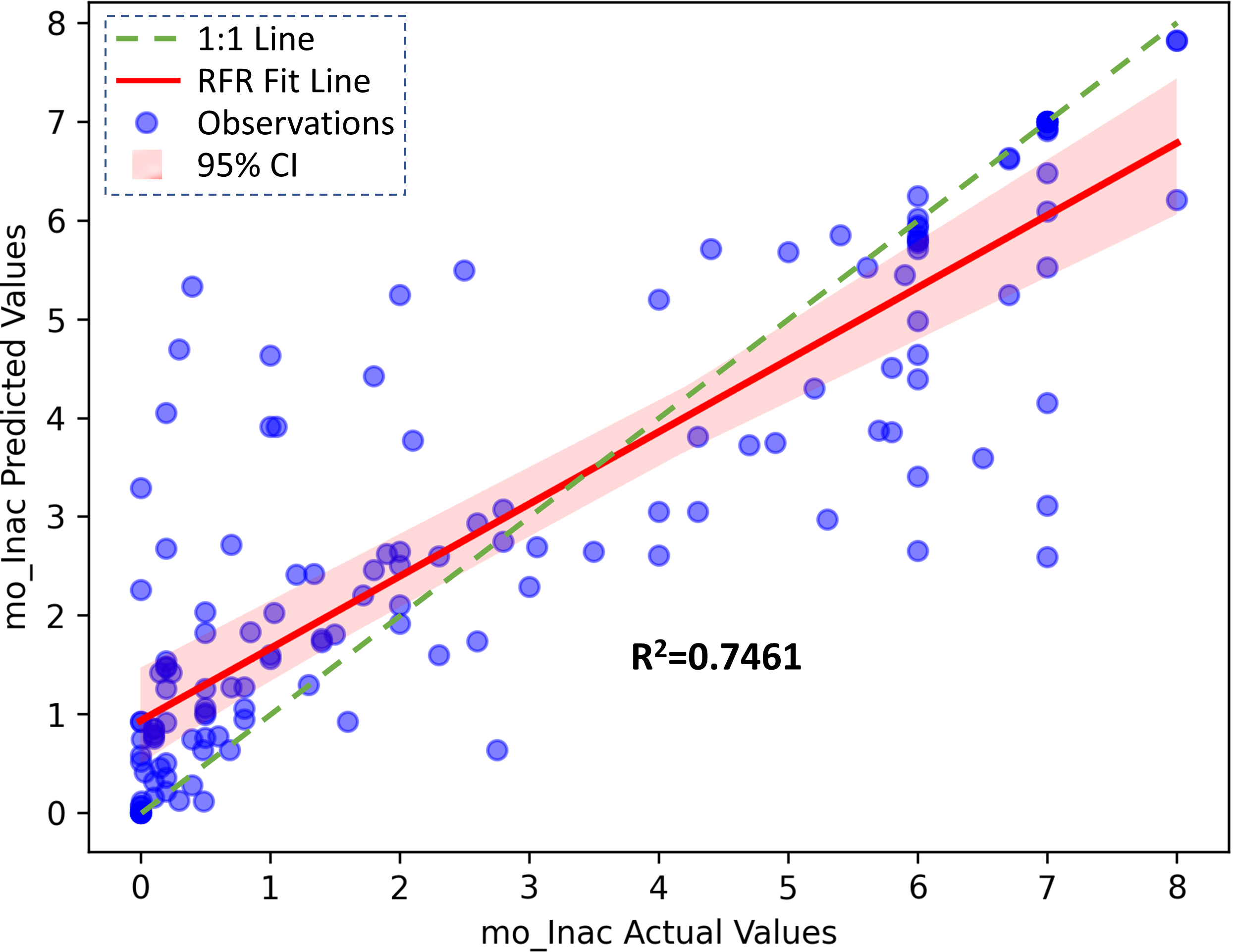}
\caption{Regression plot of the test phase (model predict) of eventual oRFR  model ($p<.001$, mo\_Inac: Microbial inactivation).}
      \label{Fig:reg_MI_graph}
\end{center}
\end{figure}

Additionally, Figure \ref{Fig:reg_MI_graph} presents the plot of predicted MI values versus actual MI values. The model's ability to predict the measured outcome (MI) variable can be examined using the graph. The residual of the fit is not directly plotted on either axis in the graph. Instead, the graph shows the predicted y value on the y-axis and the actual y value (recorded in the data table) on the x-axis. The standard diagonal line that centers the graph is the fit line for the best regressor. The vertical distance between the plotted point and the red line of identity in this instance serves as a representation of the residual (the horizontal distance can also be used as these distances will always be the same for each point). The graph demonstrates that the model performs better at predicting actual values at lower and medium Y values (between 0 and 5) which points are closer to the RFR fit line, while predictions are further off at higher Y values (points farther from the fit line). Since each data point is close to the predicted regression line, it can be concluded that the chosen RFR model fits the data quite well with the test R\textsuperscript{2} value of nearly 0.75. The graphic demonstrates the close agreement between model predictions and actual data, demonstrating the reliability of the regression model.

\begin{figure}
\begin{center}
\includegraphics[width=\textwidth,height=\textheight,keepaspectratio]{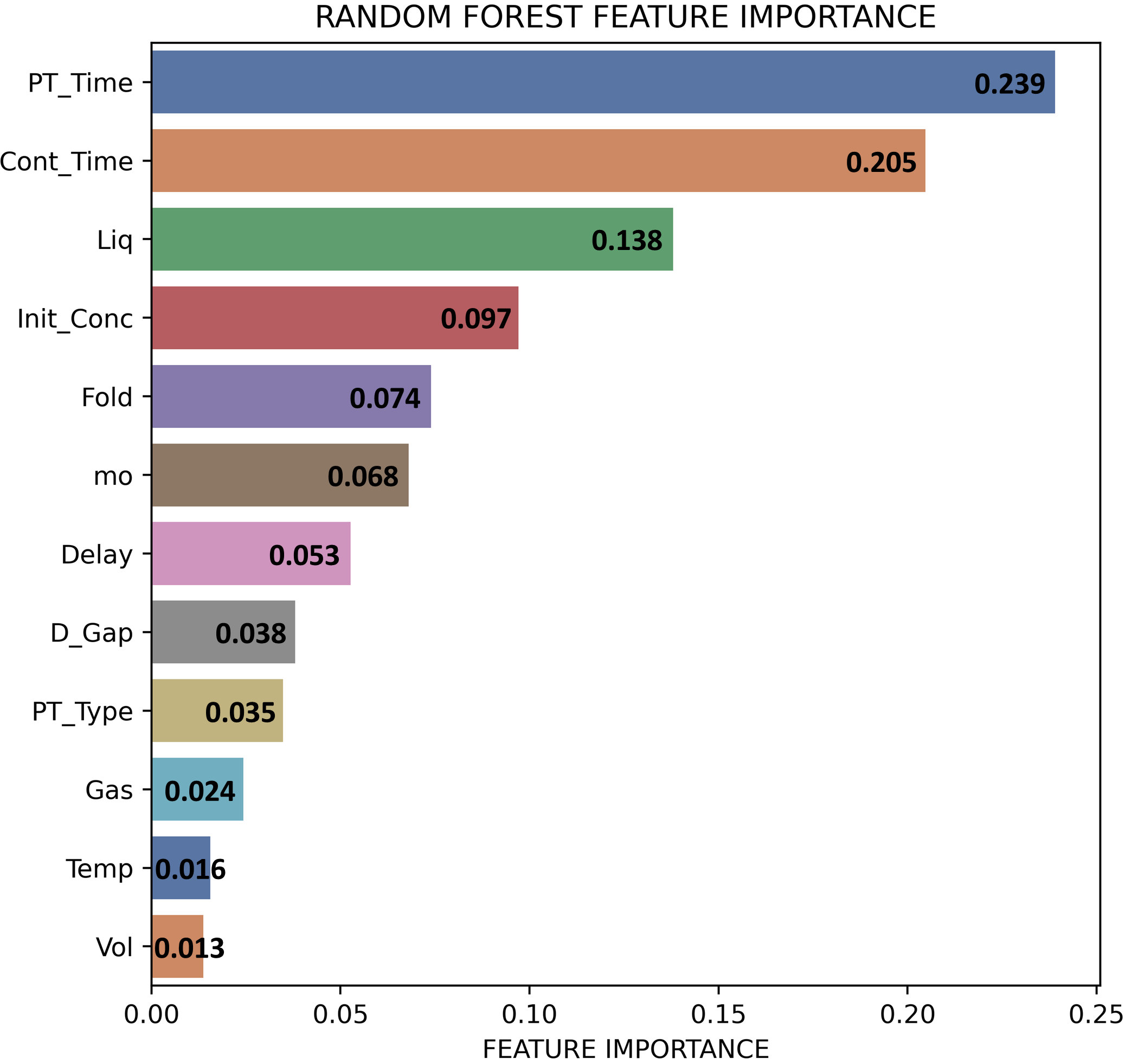}
\caption{Feature (predictor) importance results for eventual oRFR model. Feature importances were determined by considering the feature's weight wherein the final regression model function (PT\_Time: Plasma treatment time, Cont\_Time: Contact time, mo: Microbial strain, Liq: Plasma activated liquid, Delay: Post storage time, Fold: PAL/mo suspension volume ratio, Init\_Conc: Initial microbial load, D\_Gap: Discharge gap, PT\_Type: Plasma treatment type, Temp: Incubation temperature, Gas: Gas type, Vol: Treatment volume).}
      \label{Fig:reg_feature_importance}
\end{center}
\end{figure}

The objective of this study's feature importance analysis was to quantify the impact of the predictors on the output of the regression model. The scores from the feature importance analysis shed light on the dataset and model and may enhance the model's predictive performance \cite{mirzaei2021machine}. Figure \ref{Fig:reg_feature_importance} presents the outcomes of the feature importance analysis. Similar to the classification problem, feature importance analysis wasperformed using the model function's coefficients. In comparison to other predictors, plasma treatment time, contact time, liquid type, and initial microbial load were determined to be more significant features; it may be inferred that these features had the most impact on the outcome. In contrast, the regression model is less impacted by treatment volume, incubation temperature, gas type, and plasma treatment type. Compared with the feature importance graph obtained as a result of the classification model, the four features (plasma treatment type, gas type, incubation temperature, and treatment volume) that had the least impact on the output were the same as the regression model's feature importance, even though they were in different orders. Furthermore, the two most important features of both models were the contact time and plasma treatment time. When compared to other features, PAL ranked fourth in the classification model but third in the regression model. Microbial strain and initial microbial load are the only two features that have changed for the top four features in the feature importance analysis for both models. The initial concentration was the fourth important feature in the regression model, while the microbial strain was the third feature for classification. Although little differences exist, The regression features' importance was in line with the classification features importance analysis. The feature analysis results, as well as numerical data correlation results, demonstrated that all essential CAP predictors with not only various levels but also all significant might have significant impact on the building of supervised ML models.

\section{Discussion}
\label{Discussion}

In order to predict the MI of PALs, this study presented the implementation of ML in the field of plasma medicine, from data collecting through model evaluation. According to the studies in the literature, the outcomes of quantitative and qualitative analyses to determine the antimicrobial activity of PALs rely on a number of factors. The microbial strain and type of plasma-activated liquid both affect PAL's antimicrobial activity \cite{xiang2020review}. Additionally, the power, frequency, exposure period, pulse form, electrode geometry, and other parameters that affect plasma device efficiency could change the antimicrobial effect of PALs \cite{Monetta_2011}. Therefore, it is challenging to compare the antimicrobial activity of PALs produced by various devices, and optimization of parameters required to achieve a remarkable antimicrobial effect is both time-consuming and costly. Determining the plasma treatment parameters, achieving the desired antimicrobial effect, and tailoring plasma treatment to a particular standardization are significant challenges in the field of plasma medicine. The definition of ‘‘plasma dose’’ will be aided by standardization, which is both a crucial problem and a necessity in the field of plasma medicine. The issue of determining the ‘‘plasma dose’’ using Al is growing in significance despite the fact that there is numerous research for plasma dose assessment in the literature \cite{bonzanini2021perspectives,Sakai_2022}. In this study, it is expected that the widespread use of Al in the medical and biomedical fields today will enable the targeted standardization for plasma medicine. To the best of our knowledge, no studies have previously employed ML models to predict the antimicrobial activity of PALs. Some research in the literature claims that PALs' antimicrobial activity is significantly influenced by a number of different factors. Plasma treatment time, contact time, liquid type, microbial strain, and PAL/mo suspension volume ratio are a few examples of these characteristics \cite{kojtari2013chemistry,zhao2020inactivation,tsoukou2018understanding,kamgang2008impact}. This study's prediction model is based on the different research groups' features that are examining the antimicrobial efficacy of PALs.

In this study, ML was based on various preprocessing steps as well as different classification or regression models due to the select best combination for building and training the models. Also, obtained results for training, validation, and testing phases were evaluated in a comprehensive and fair way. Therefore, many classification and regression scenarios, parameter tuning approaches, cross-validation strategies, and statistical analyses were carried out and robust statistical metrics were provided. For this direction, different normalizing strategies were utilzed to provide data preparation for model training. The models trained using the original data before the normalization techniques have shown inferior prediction accuracy R\textsuperscript{2} scores. The main cause of this is that the created dataset is wide unit ranges and before training the model, normalization techniques limit the ranges of the dataset for each feature to make model training easier. For classification and regression models, four alternative normalization techniques were tested. The precision capabilities of both models improved after normalization, however, while the Z-score normalization method revealed better accuracy for classification, the Robust scaler method yielded better performance for regression models by using the same feature set.

In ML, the high accuracy of the created model is very important both for the usage and for evaluating its functionality in terms of realization of the target and suitability for real life. Especially, the high precision of ML studies aimed to be used in the field of health is a very necessary and desirable feature \cite{fleuren2020machine}. As mentioned before, there are only a few studies reported in the literature for the prediction of different biological outcomes by using ML techniques and they have a great impact on their fields due to the advantages of decreasing the experimental cost and time consumption. Nonetheless, their prediction performance is limited. For instance, Shaban et al. presented prediction accuracy of 67\% ±6.1\% for the regression model, 73\% ±5.8\% for the classification model, and 74\% ±5\% for the random forest model, as a result of the prediction of the antibiofilm activity of antibiotics using supervised machine learning \cite{shaban2022prediction}. In another study, Mirzaei et al. created a predictive model of the antibacterial activity of nanoparticles and demonstrated a regression model with the R\textsuperscript{2} value of 0.78 \cite{mirzaei2021machine}. In another study \cite{furxhi2020predicting}, the RF model was implemented to predict the neurotoxicity of nanoparticles via ML. As a result of the study, the ACC score was achieved as 72\%. Overall, when other results of similar studies are taken into account and compared, the accuracy and R\textsuperscript{2} values outperformed those presented in this study and both of the created supervised ML models are more promising than other studies when considering the prediction performance.

The results of this study revealed that oRFC and oRFR were the most effective models for classification and regression. Furthermore, no overfitting has occurred for any supervised ML models. Due to their efficient operation on highly dimensional data and their excellent accuracy, RF models are a prominent data analysis tool in research for medical and biomedical applications \cite{garg2021role}. RF models are typically applied to high-dimensional data sets without a linear relationship between the variables used as inputs and outputs \cite{li2021effective}. Because the characterization of the data used in this study is high-dimensional and non-linear it is appropriate for the RF model. In line with the literature,  in this study, encouraging performances were obtained compared to other classification and regression algorithms by using RF-based models. Also, the RF-based models which are structured on ensemble learning and trees might have been superior performance to other models because of the data distribution.

Additionally, when the test data were predicted with trained models, the oRFC yielded an accuracy value of 82.68\%, whereas oRFR achieved R\textsuperscript{2} value of 0.75. For the comparison of two different supervised ML techniques, the best classification model (oRFC), in which the output variable (MI value) is categorized, outperformed the regression model (oRFR), in which the MI value is provided as a continuous variable. Regression models' prediction capability may be less than that of classification models because regression models attempt to estimate the exact MI values and MI has a wide distribution. For the field of plasma medicine, it may be more crucial to estimate the output value as the exact value when MI is considered as a biological output, but the model's prediction ability has been enhanced by the categorizing strategy utilized in the study. It is obvious that there is a trade-off between the ability to predict outcomes and the categorization of the MI variable. Additionally, both regression models using MI as continuous variable and classification models using MI as a categorical variable have both been developed and discussed for usage. Classification or regression techniques may be preferred for the desired task by considering the advantages and disadvantages.

On the other hand, according to state-of-the-art studies, the plasma treatment time \cite{kojtari2013chemistry}, contact time \cite{zhao2020inactivation}, liquid type \cite{tsoukou2018understanding}, and initial microorganism concentration \cite{kamgang2008impact} are the variables that have the biggest impacts on the antimicrobial efficacy of PALs. They were picked as part of the study's data acquisition parameters as a consequence. Among all predictors, the plasma treatment time, contact time, and PAL type variables were shown to be the most significant features for obtaining a remarkable antimicrobial activity of PALs and highly impactful on created models on the results of feature importance analysis. These three variables, which were found to be the most important ones among other predictors, are also consistent with the literature. It is well known that antimicrobial activity directly correlates with the plasma treatment time. Depending on the type and volume of PAL, it has been reported in the literature that the effectiveness of MI increases as the time of treatment of the liquid with CAP increases \cite{kaushik2019biological,xiang2022review}. MI is influenced by the length of time the generated PAL is in contact with the bacterial solution. During the period of contact, the plasma content that was transferred from the CAP to the liquid participates in a number of biochemical processes with microorganisms. Although the ideal duration for these reactions should be standardized, in some liquids it is vital to do so in order to prevent the plasma species' scavenging impact \cite{schmidt2019plasma}. An extremely useful indicator of antimicrobial activity is the liquid form of PAL generated for a particular antimicrobial application. Changing the type of liquid also affects the liquid's chemical composition. It has been found that the molecules in this chemical structure interact with the reactive species in the plasma's active component to create new molecules \cite{verlackt2018transport}. For classification models, the fourth important feature was the microbial strain and for regression models, it was the initial microbial load. Both of them are related to in vitro characteristics. Also, both the microbial strain and initial microbial load features are crucial when comparing and evaluating the antimicrobial activity of PALs that are designed to target study \cite{lee2019antibacterial}. Studies in the literature indicate that these characteristics, which are essential to the antimicrobial efficacy of PALs, correspond to the parameters identified by the trained model's feature importance analysis. This correlation demonstrated how well the training model fit and predicted the experimental data. Moreover, the classification model has been rebuilt with the four most important features to see whether the accuracy of the model has changed. The oRFC model yielded a 65.80\% accuracy value by using only the top four important predictors which were determined by the feature importance analysis. Furthermore, it was decided that testing the remaining 8 features was necessary because the accuracy of the model built with the most important four features was quite low. Therefore, the oRFC model was rebuilt with the remaining 8 features to observe the accuracy changing effect and achieved an accuracy value of 58.87\%. Both rebuilt models' accuracy values were far below from the eventual test result of 82.68\%. The rebuilt model findings demonstrate that all predictors have an impact on the model, even though the features with the largest and lowest contributions to the model were noted following feature importance analysis.

Despite the encouraging results of the present study, a number of limitations should be considered. First, the trained model was created using the data from the articles that were used. The data from the research in the literature were the only sources on which the prediction model was built. The models' accuracy and \textsuperscript{2} score were unaffected by this condition, but it prevented the model from being validated using actual data. Because the accuracy of the studies that were performed is a requirement for the prediction model that was presented in this study, this constraint needs to be removed. After the model was developed, it should be tested by obtaining real data and comparing the prediction results to the experimental validation. The lack of observations is another limitation of this study. The amount of data used for this study should be increased and the effect of the observation variance on the prediction ability of ML models should be decreased in this way. Higher data size in ML denotes the model's robustness. In order to increase model validation, a dataset with a lot more observations is required. Briefly, a higher dimensional dataset and in vitro tests are required to more accurately examine the antimicrobial effectiveness of PALs.

AI is effective in several biomedical fields, which encourages its application in plasma medicine. This study is the first to show how ML modeling can qualitatively predict the antimicrobial activity of PALs in the field of plasma medicine. This research is a pioneering study for the future development of qualitative and quantitative prediction models in plasma medicine applications via ML. The study's final results are promising and also capable of automatically predicting the MI for both models. The handling of input data which has diverse distributions by ML, one of the subfields of AI, also demonstrated remarkable performance in terms of computational cost and prediction capability. The results demonstrated that an adroit combination of ML techniques with CAP-related data might have a significant impact on plasma medicine besides considering that AI studies are wide usage and of crucial importance in other biomedical research fields. A comparison of real-time experimental results with prediction model results will demonstrate the applicability of ML in plasma medicine and demonstrate how it relates to daily life. In conclusion, an ML model that can automatically predict the MI value with high prediction ability has been created. It should be also noted the fact that experimental studies are expensive, time-consuming, and difficult to show the experimental outcome. The obtained findings encourage the potential for merging ML techniques with plasma medicine applications to determine the ‘‘plasma dose’’ as well as adopting AI approaches to the prediction of other important parameters used in plasma medicine in the future. Also, the results are in line with the literature on the manner of biological aspects of CAP.

\section{Conclusions}
\label{Conclusions}

In conclusion, in the present study, we have utilized the ML techniques to predict the antimicrobial activity of the PALs which depends on the various plasma treatment parameters and hard to make comparisons in between the different studies. We conducted comprehensive and various robust training and test scenarios, as well as statistical analysis, for building and testing supervised ML models to evaluate fairly the results. Our results revealed that the generated models in the present study can predict the antimicrobial activity of PALs with a test accuracy of 82.68\% for the categorical outcome and with a test R\textsuperscript{2} score of 0.75 for exact values of outcome based on the available literature. Also, RF-based models demonstrated superiority among other algorithms for the prediction of MI in both regression and classification problems. Furthermore, the importance of features that was determined through the model was in line with the literature where many studies have shown the plasma treatment time as one of the primary parameters for the antimicrobial effect of PALs. Best of our knowledge this study is the first to utilize ML techniques in the field of plasma medicine. By considering the wide range of applications of CAP in medicine and biology, ML techniques may assist in better understanding the biological outcomes of CAP applications. Furthermore, ML applications in plasma medicine may contribute to defining the ‘‘plasma dose’’ that is a contemporary concept in the field of plasma medicine and thought to be a crucial concept to convey various CAP applications into clinical practice. 

\section*{CRediT authorship contribution statement}
\textbf{Mehmet Akif Özdemir:} Conceptualization, Methodology, Software, Validation, Investigation, Visualization, Writing-Original Draft, Writing-Review {\&} Editing.
\textbf{Gizem Dilara Özdemir:} Conceptualization, Methodology, Investigation,  Visualization, Data Curation, Formal Analysis, Writing-Original Draft, Writing-Review {\&} Editing.
\textbf{Merve Gül:} Data Curation, Writing-Original Draft.
\textbf{Onan Güren:} Conceptualization, Writing-Review {\&} Editing.
\textbf{Utku Kürşat Ercan:} Conceptualization, Methodology, Funding Acquisition, Writing-Original Draft, Writing-Review {\&} Editing, Supervision. All authors read and approved the final manuscript.

\section*{Declaration of Competing Interest}

The authors declare that they have no known competing financial interests or personal relationships that could have appeared to influence the work reported in this paper.

\section*{Acknowledgements}

This work was supported by the Izmir Katip Celebi University Scientific Research Projects Coordination Unit [grant number 2022-ÖDL-MÜMF-0004].

\section*{Ethics and Participation Approvals}

Not applicable.

\section*{Data Statement}

The data that support the findings of this study are available upon request.

\bibliographystyle{elsarticle-num.bst}
\bibliography{references}

\begin{thebibliography}{10}
\expandafter\ifx\csname url\endcsname\relax
  \def\url#1{\texttt{#1}}\fi
\expandafter\ifx\csname urlprefix\endcsname\relax\def\urlprefix{URL }\fi
\expandafter\ifx\csname href\endcsname\relax
  \def\href#1#2{#2} \def\path#1{#1}\fi

\bibitem{moreau2008non}
M.~Moreau, N.~Orange, M.~Feuilloley, Non-thermal plasma technologies: new tools
  for bio-decontamination, Biotechnology advances 26~(6) (2008) 610--617.
\newblock \href {https://doi.org/10.1016/j.biotechadv.2008.08.001}
  {\path{doi:10.1016/j.biotechadv.2008.08.001}}.

\bibitem{metelmann2018comprehensive}
H.-R. Metelmann, T.~Von~Woedtke, K.-D. Weltmann, Comprehensive clinical plasma
  medicine: cold physical plasma for medical application, Springer, 2018.

\bibitem{oztan2022irrigation}
M.~O. Oztan, U.~K. Ercan, A.~Aksoy~Gokmen, F.~Simsek, G.~D. Ozdemir,
  G.~Koyluoglu, Irrigation of peritoneal cavity with cold atmospheric plasma
  treated solution effectively reduces microbial load in rat acute peritonitis
  model, Scientific Reports 12~(1) (2022) 1--15.
\newblock \href {https://doi.org/10.1038/s41598-022-07598-2}
  {\path{doi:10.1038/s41598-022-07598-2}}.

\bibitem{kim2021applications}
S.~Kim, C.-H. Kim, Applications of plasma-activated liquid in the medical
  field, Biomedicines 9~(11) (2021) 1700.
\newblock \href {https://doi.org/10.3390/biomedicines9111700}
  {\path{doi:10.3390/biomedicines9111700}}.

\bibitem{kaushik2019biological}
N.~K. Kaushik, B.~Ghimire, Y.~Li, M.~Adhikari, M.~Veerana, N.~Kaushik, N.~Jha,
  B.~Adhikari, S.-J. Lee, K.~Masur, et~al., Biological and medical applications
  of plasma-activated media, water and solutions, Biological chemistry 400~(1)
  (2019) 39--62.
\newblock \href {https://doi.org/doi:10.1515/hsz-2018-0226}
  {\path{doi:doi:10.1515/hsz-2018-0226}}.

\bibitem{tsoukou2018understanding}
E.~Tsoukou, P.~Bourke, D.~Boehm, Understanding the differences between
  antimicrobial and cytotoxic properties of plasma activated liquids, Plasma
  Medicine 8~(3) (2018) 299--320.
\newblock \href {https://doi.org/10.1615/PlasmaMed.2018028261}
  {\path{doi:10.1615/PlasmaMed.2018028261}}.

\bibitem{xiang2020review}
Q.~Xiang, L.~Fan, Y.~Li, S.~Dong, K.~Li, Y.~Bai, A review on recent advances in
  plasma-activated water for food safety: Current applications and future
  trends, Critical Reviews in Food Science and Nutrition 62~(8) (2022)
  2250--2268.
\newblock \href {https://doi.org/10.1080/10408398.2020.1852173}
  {\path{doi:10.1080/10408398.2020.1852173}}.

\bibitem{ercan2013nonequilibrium}
U.~K. Ercan, H.~Wang, H.~Ji, G.~Fridman, A.~D. Brooks, S.~G. Joshi,
  Nonequilibrium plasma-activated antimicrobial solutions are broad-spectrum
  and retain their efficacies for extended period of time, Plasma processes and
  polymers 10~(6) (2013) 544--555.
\newblock \href {https://doi.org/https://doi.org/10.1002/ppap.201200104}
  {\path{doi:https://doi.org/10.1002/ppap.201200104}}.

\bibitem{schmidt2019plasma}
M.~Schmidt, V.~Hahn, B.~Altrock, T.~Gerling, I.~C. Gerber, K.-D. Weltmann,
  T.~von Woedtke, Plasma-activation of larger liquid volumes by an
  inductively-limited discharge for antimicrobial purposes, Applied Sciences
  9~(10) (2019) 2150.
\newblock \href {https://doi.org/10.3390/app9102150}
  {\path{doi:10.3390/app9102150}}.

\bibitem{cheng2019colorimetric}
J.~Cheng, Q.~Chen, G.~Fridman, H.-F. Ji, A colorimetric method for comparison
  of oxidative strength of dbd plasma, Sensors and Actuators Reports 1 (2019)
  100001.
\newblock \href {https://doi.org/10.1016/j.snr.2019.100001}
  {\path{doi:10.1016/j.snr.2019.100001}}.

\bibitem{mirzaei2021machine}
M.~Mirzaei, I.~Furxhi, F.~Murphy, M.~Mullins, A machine learning tool to
  predict the antibacterial capacity of nanoparticles, Nanomaterials 11~(7)
  (2021) 1774.
\newblock \href {https://doi.org/10.3390/nano11071774}
  {\path{doi:10.3390/nano11071774}}.

\bibitem{rajkomar2019machine}
A.~Rajkomar, J.~Dean, I.~Kohane, Machine learning in medicine, New England
  Journal of Medicine 380~(14) (2019) 1347--1358.
\newblock \href {https://doi.org/10.1056/NEJMra1814259}
  {\path{doi:10.1056/NEJMra1814259}}.

\bibitem{furxhi2019machine}
I.~Furxhi, F.~Murphy, M.~Mullins, C.~A. Poland, Machine learning prediction of
  nanoparticle in vitro toxicity: A comparative study of classifiers and
  ensemble-classifiers using the copeland index, Toxicology letters 312 (2019)
  157--166.
\newblock \href {https://doi.org/10.1016/j.toxlet.2019.05.016}
  {\path{doi:10.1016/j.toxlet.2019.05.016}}.

\bibitem{serafim2020application}
M.~S.~M. Serafim, T.~Kronenberger, P.~R. Oliveira, A.~Poso, K.~M. Honorio,
  B.~E.~F. Mota, V.~G. Maltarollo, The application of machine learning
  techniques to innovative antibacterial discovery and development, Expert
  opinion on drug discovery 15~(10) (2020) 1165--1180.
\newblock \href {https://doi.org/10.1080/17460441.2020.1776696}
  {\path{doi:10.1080/17460441.2020.1776696}}.

\bibitem{liu2020evaluation}
Z.~Liu, D.~Deng, H.~Lu, J.~Sun, L.~Lv, S.~Li, G.~Peng, X.~Ma, J.~Li, Z.~Li,
  et~al., Evaluation of machine learning models for predicting antimicrobial
  resistance of actinobacillus pleuropneumoniae from whole genome sequences,
  Frontiers in microbiology 11 (2020) 48.
\newblock \href {https://doi.org/10.3389/fmicb.2020.00048}
  {\path{doi:10.3389/fmicb.2020.00048}}.

\bibitem{shaban2022prediction}
T.~F. Shaban, M.~Y. Alkawareek, Prediction of qualitative antibiofilm activity
  of antibiotics using supervised machine learning techniques, Computers in
  biology and medicine 140 (2022) 105065.
\newblock \href {https://doi.org/10.1016/j.compbiomed.2021.105065}
  {\path{doi:10.1016/j.compbiomed.2021.105065}}.

\bibitem{furxhi2020practices}
I.~Furxhi, F.~Murphy, M.~Mullins, A.~Arvanitis, C.~A. Poland, Practices and
  trends of machine learning application in nanotoxicology, Nanomaterials
  10~(1) (2020) 116.
\newblock \href {https://doi.org/10.3390/nano10010116}
  {\path{doi:10.3390/nano10010116}}.

\bibitem{furxhi2020predicting}
I.~Furxhi, F.~Murphy, Predicting in vitro neurotoxicity induced by
  nanoparticles using machine learning, International journal of molecular
  sciences 21~(15) (2020) 5280.
\newblock \href {https://doi.org/10.3390/ijms21155280}
  {\path{doi:10.3390/ijms21155280}}.

\bibitem{kojtari2013chemistry}
A.~Kojtari, U.~Ercan, J.~Smith, G.~Friedman, R.~Sensenig, S.~Tyagi, S.~Joshi,
  H.~Ji, A.~Brooks, Chemistry for antimicrobial properties of water treated
  with non-equilibrium plasma, J. Nanomed. Biotherapeutic Discovery 4~(1)
  (2013) 120.
\newblock \href {https://doi.org/10.4172/2155-983X.1000120}
  {\path{doi:10.4172/2155-983X.1000120}}.

\bibitem{hong2021antimicrobial}
Q.~Hong, X.~Dong, H.~Yu, H.~Sun, M.~Chen, Y.~Wang, Q.~Yu, The antimicrobial
  property of plasma activated liquids (pals) against oral bacteria
  streptococcus mutans, Dental 3~(1) (2021) 1--7.
\newblock \href {https://doi.org/10.35702/dent.10007}
  {\path{doi:10.35702/dent.10007}}.

\bibitem{smet2019inactivation}
C.~Smet, M.~Govaert, A.~Kyrylenko, M.~Easdani, J.~L. Walsh, J.~F. Van~Impe,
  Inactivation of single strains of listeria monocytogenes and salmonella
  typhimurium planktonic cells biofilms with plasma activated liquids,
  Frontiers in microbiology 10 (2019) 1539.
\newblock \href {https://doi.org/10.3389/fmicb.2019.01539}
  {\path{doi:10.3389/fmicb.2019.01539}}.

\bibitem{shen2016bactericidal}
J.~Shen, Y.~Tian, Y.~Li, R.~Ma, Q.~Zhang, J.~Zhang, J.~Fang, Bactericidal
  effects against s. aureus and physicochemical properties of plasma activated
  water stored at different temperatures, Scientific reports 6~(1) (2016)
  1--10.
\newblock \href {https://doi.org/10.1038/srep28505}
  {\path{doi:10.1038/srep28505}}.

\bibitem{zhao2020inactivation}
Y.-M. Zhao, S.~Ojha, C.~Burgess, D.-W. Sun, B.~Tiwari, Inactivation efficacy
  and mechanisms of plasma activated water on bacteria in planktonic state,
  Journal of Applied Microbiology 129~(5) (2020) 1248--1260.
\newblock \href {https://doi.org/10.1111/jam.14677}
  {\path{doi:10.1111/jam.14677}}.

\bibitem{li2017vitro}
Y.~Li, J.~Pan, G.~Ye, Q.~Zhang, J.~Wang, J.~Zhang, J.~Fang, In vitro studies of
  the antimicrobial effect of non-thermal plasma-activated water as a novel
  mouthwash, European Journal of Oral Sciences 125~(6) (2017) 463--470.
\newblock \href {https://doi.org/10.1111/eos.12374}
  {\path{doi:10.1111/eos.12374}}.

\bibitem{liu2021vitro}
J.~Liu, C.~Yang, C.~Cheng, C.~Zhang, J.~Zhao, C.~Fu, In vitro antimicrobial
  effect and mechanism of action of plasma-activated liquid on planktonic
  neisseria gonorrhoeae, Bioengineered 12~(1) (2021) 4605--4619.
\newblock \href {https://doi.org/10.1080/21655979.2021.1955548}
  {\path{doi:10.1080/21655979.2021.1955548}}.

\bibitem{oehmigen2011estimation}
K.~Oehmigen, J.~Winter, M.~H{\"a}hnel, C.~Wilke, R.~Brandenburg, K.-D.
  Weltmann, T.~von Woedtke, Estimation of possible mechanisms of escherichia
  coli inactivation by plasma treated sodium chloride solution, Plasma
  Processes and Polymers 8~(10) (2011) 904--913.
\newblock \href {https://doi.org/.1002/ppap.201000099}
  {\path{doi:.1002/ppap.201000099}}.

\bibitem{kamgang2008impact}
G.~Kamgang-Youbi, J.-M. Herry, J.-L. Brisset, M.-N. Bellon-Fontaine, A.~Doubla,
  M.~Na{\"\i}tali, Impact on disinfection efficiency of cell load and of
  planktonic/adherent/detached state: case of hafnia alvei inactivation by
  plasma activated water, Applied microbiology and biotechnology 81~(3) (2008)
  449--457.
\newblock \href {https://doi.org/10.1007/s00253-008-1641-9}
  {\path{doi:10.1007/s00253-008-1641-9}}.

\bibitem{kamgang2009microbial}
G.~Kamgang-Youbi, J.-M. Herry, T.~Meylheuc, J.-L. Brisset, M.-N.
  Bellon-Fontaine, A.~Doubla, M.~Naitali, Microbial inactivation using
  plasma-activated water obtained by gliding electric discharges, Letters in
  applied microbiology 48~(1) (2009) 13--18.
\newblock \href {https://doi.org/10.1111/j.1472-765X.2008.02476.x}
  {\path{doi:10.1111/j.1472-765X.2008.02476.x}}.

\bibitem{dezest2017oxidative}
M.~Dezest, A.-L. Bulteau, D.~Quinton, L.~Chavatte, M.~Le~B{\'e}chec, J.~P.
  Cambus, S.~Arbault, A.~N{\`e}gre-Salvayre, F.~Clement, S.~Cousty, Oxidative
  modification and electrochemical inactivation of escherichia coli upon cold
  atmospheric pressure plasma exposure, PLoS One 12~(3) (2017) e0173618.
\newblock \href {https://doi.org/10.1371/journal.pone.0173618}
  {\path{doi:10.1371/journal.pone.0173618}}.

\bibitem{simon2022influence}
S.~Simon, B.~Salgado, M.~I. Hasan, M.~Sivertsvik, E.~N. Fern{\'a}ndez, J.~L.
  Walsh, Influence of potable water origin on the physicochemical and
  antimicrobial properties of plasma activated water, Plasma Chemistry and
  Plasma Processing 42~(2) (2022) 377--393.
\newblock \href {https://doi.org/10.1007/s11090-021-10221-3}
  {\path{doi:10.1007/s11090-021-10221-3}}.

\bibitem{zhou2018cold}
R.~Zhou, R.~Zhou, K.~Prasad, Z.~Fang, R.~Speight, K.~Bazaka, K.~K. Ostrikov,
  Cold atmospheric plasma activated water as a prospective disinfectant: the
  crucial role of peroxynitrite, Green Chemistry 20~(23) (2018) 5276--5284.
\newblock \href {https://doi.org/10.1039/C8GC02800A}
  {\path{doi:10.1039/C8GC02800A}}.

\bibitem{chiappim2021antimicrobial}
W.~Chiappim, A.~d.~G. Sampaio, F.~Miranda, M.~Fraga, G.~Petraconi,
  A.~da~Silva~Sobrinho, K.~Kostov, C.~Koga-Ito, R.~Pessoa, Antimicrobial effect
  of plasma-activated tap water on staphylococcus aureus, escherichia coli, and
  candida albicans, Water 13~(11) (2021) 1480.
\newblock \href {https://doi.org/10.3390/w13111480}
  {\path{doi:10.3390/w13111480}}.

\bibitem{hansch2015analysis}
M.~A. H{\"a}nsch, M.~Mann, K.-D. Weltmann, T.~Von~Woedtke, Analysis of
  antibacterial efficacy of plasma-treated sodium chloride solutions, Journal
  of Physics D: Applied Physics 48~(45) (2015) 454001.
\newblock \href {https://doi.org/10.1088/0022-3727/48/45/454001}
  {\path{doi:10.1088/0022-3727/48/45/454001}}.

\bibitem{rathore2021investigation}
V.~Rathore, D.~Patel, S.~Butani, S.~K. Nema, Investigation of physicochemical
  properties of plasma activated water and its bactericidal efficacy, Plasma
  Chemistry and Plasma Processing 41~(3) (2021) 871--902.
\newblock \href {https://doi.org/10.1007/s11090-021-10161-y}
  {\path{doi:10.1007/s11090-021-10161-y}}.

\bibitem{joshi2018characterization}
I.~Joshi, D.~Salvi, D.~W. Schaffner, M.~V. Karwe, Characterization of microbial
  inactivation using plasma-activated water and plasma-activated acidified
  buffer, Journal of food protection 81~(9) (2018) 1472--1480.
\newblock \href {https://doi.org/10.4315/0362-028X.JFP-17-487}
  {\path{doi:10.4315/0362-028X.JFP-17-487}}.

\bibitem{ma2020key}
M.~Ma, Y.~Zhang, Y.~Lv, F.~Sun, The key reactive species in the bactericidal
  process of plasma activated water, Journal of Physics D: Applied Physics
  53~(18) (2020) 185207.
\newblock \href {https://doi.org/10.1088/1361-6463/ab703a}
  {\path{doi:10.1088/1361-6463/ab703a}}.

\bibitem{naitali2010combined}
M.~Na{\"\i}tali, G.~Kamgang-Youbi, J.-M. Herry, M.-N. Bellon-Fontaine, J.-L.
  Brisset, Combined effects of long-living chemical species during microbial
  inactivation using atmospheric plasma-treated water, Applied and
  environmental microbiology 76~(22) (2010) 7662--7664.
\newblock \href {https://doi.org/10.1128/AEM.01615-10}
  {\path{doi:10.1128/AEM.01615-10}}.

\bibitem{qi2018inactivation}
Z.~Qi, E.~Tian, Y.~Song, E.~A. Sosnin, V.~S. Skakun, T.~Li, Y.~Xia, Y.~Zhao,
  X.~Lin, D.~Liu, Inactivation of shewanella putrefaciens by plasma activated
  water, Plasma chemistry and plasma processing 38~(5) (2018) 1035--1050.
\newblock \href {https://doi.org/10.1007/s11090-018-9911-5}
  {\path{doi:10.1007/s11090-018-9911-5}}.

\bibitem{royintarat2019mechanism}
T.~Royintarat, P.~Seesuriyachan, D.~Boonyawan, E.~H. Choi, W.~Wattanutchariya,
  Mechanism and optimization of non-thermal plasma-activated water for
  bacterial inactivation by underwater plasma jet and delivery of reactive
  species underwater by cylindrical dbd plasma, Current Applied Physics 19~(9)
  (2019) 1006--1014.
\newblock \href {https://doi.org/10.1016/j.cap.2019.05.020}
  {\path{doi:10.1016/j.cap.2019.05.020}}.

\bibitem{suganuma2018air}
R.~Suganuma, K.~Yasuoka, Air-supplied pinhole discharge in aqueous solution for
  the inactivation of escherichia coli, Japanese Journal of Applied Physics
  57~(4) (2018) 046202.
\newblock \href {https://doi.org/10.7567/jjap.57.046202}
  {\path{doi:10.7567/jjap.57.046202}}.

\bibitem{tian2015assessment}
Y.~Tian, R.~Ma, Q.~Zhang, H.~Feng, Y.~Liang, J.~Zhang, J.~Fang, Assessment of
  the physicochemical properties and biological effects of water activated by
  non-thermal plasma above and beneath the water surface, Plasma processes and
  polymers 12~(5) (2015) 439--449.
\newblock \href {https://doi.org/10.1002/ppap.201400082}
  {\path{doi:10.1002/ppap.201400082}}.

\bibitem{traylor2011long}
M.~J. Traylor, M.~J. Pavlovich, S.~Karim, P.~Hait, Y.~Sakiyama, D.~S. Clark,
  D.~B. Graves, Long-term antibacterial efficacy of air plasma-activated water,
  Journal of Physics D: Applied Physics 44~(47) (2011) 472001.
\newblock \href {https://doi.org/10.1088/0022-3727/44/47/472001}
  {\path{doi:10.1088/0022-3727/44/47/472001}}.

\bibitem{wu2017reactive}
S.~Wu, Q.~Zhang, R.~Ma, S.~Yu, K.~Wang, J.~Zhang, J.~Fang, Reactive
  radical-driven bacterial inactivation by hydrogen-peroxide-enhanced
  plasma-activated-water, The European Physical Journal Special Topics 226~(13)
  (2017) 2887--2899.
\newblock \href {https://doi.org/10.1140/epjst/e2016-60330-y}
  {\path{doi:10.1140/epjst/e2016-60330-y}}.

\bibitem{xiang2019synergistic}
Q.~Xiang, W.~Wang, D.~Zhao, L.~Niu, K.~Li, Y.~Bai, Synergistic inactivation of
  escherichia coli o157: H7 by plasma-activated water and mild heat, Food
  Control 106 (2019) 106741.
\newblock \href {https://doi.org/10.1016/j.foodcont.2019.106741}
  {\path{doi:10.1016/j.foodcont.2019.106741}}.

\bibitem{Ye2013EfficiencyAM}
G.~Ye, Q.~Zhang, H.~Pan, G.~Wang, Y.~Li, J.~Pan, J.~Wang, J.~Zhang, J.~Fang,
  Efficiency and mechanism of pathogenic bacteria inactivation by non-thermal
  plasma activated water, in: Proceedings of the 21st International Symposium
  on Plasma Chemistry, 2013, p.~1.

\bibitem{zhang2013study}
Q.~Zhang, Y.~Liang, H.~Feng, R.~Ma, Y.~Tian, J.~Zhang, J.~Fang, A study of
  oxidative stress induced by non-thermal plasma-activated water for bacterial
  damage, Applied physics letters 102~(20) (2013) 203701.
\newblock \href {https://doi.org/10.1063/1.4807133}
  {\path{doi:10.1063/1.4807133}}.

\bibitem{zhang2016sterilization}
Q.~Zhang, R.~Ma, Y.~Tian, B.~Su, K.~Wang, S.~Yu, J.~Zhang, J.~Fang,
  Sterilization efficiency of a novel electrochemical disinfectant against
  staphylococcus aureus, Environmental Science \& Technology 50~(6) (2016)
  3184--3192.
\newblock \href {https://doi.org/10.1021/acs.est.5b05108}
  {\path{doi:10.1021/acs.est.5b05108}}.

\bibitem{joshi2015microarray}
S.~G. Joshi, A.~Yost, S.~S. Joshi, S.~Addya, G.~Ehrlich, A.~Brooks, et~al.,
  Microarray analysis of transcriptomic response of escherichia coli to
  nonthermal plasma-treated pbs solution, Advances in Bioscience and
  Biotechnology 6~(02) (2015) 49.
\newblock \href {https://doi.org/10.4236/abb.2015.62006}
  {\path{doi:10.4236/abb.2015.62006}}.

\bibitem{ksikazek2019novel}
W.~Ksiazek, M.~Abdar, U.~R. Acharya, P.~P{\l}awiak, A novel machine learning
  approach for early detection of hepatocellular carcinoma patients, Cognitive
  Systems Research 54 (2019) 116--127.
\newblock \href {https://doi.org/10.1016/j.cogsys.2018.12.001}
  {\path{doi:10.1016/j.cogsys.2018.12.001}}.

\bibitem{pandey2017comparative}
A.~Pandey, A.~Jain, Comparative analysis of knn algorithm using various
  normalization techniques, International Journal of Computer Network and
  Information Security 9~(11) (2017) 36.
\newblock \href {https://doi.org/10.5815/ijcnis.2017.11.04}
  {\path{doi:10.5815/ijcnis.2017.11.04}}.

\bibitem{ahsan2021effect}
M.~M. Ahsan, M.~P. Mahmud, P.~K. Saha, K.~D. Gupta, Z.~Siddique, Effect of data
  scaling methods on machine learning algorithms and model performance,
  Technologies 9~(3) (2021) 52.
\newblock \href {https://doi.org/10.3390/technologies9030052}
  {\path{doi:10.3390/technologies9030052}}.

\bibitem{yang2019identifying}
X.~Yang, Y.~Gong, N.~Waheed, K.~March, J.~Bian, W.~R. Hogan, Y.~Wu, Identifying
  cancer patients at risk for heart failure using machine learning methods, in:
  AMIA Annual Symposium Proceedings, Vol. 2019, American Medical Informatics
  Association, 2019, pp. 933--941.

\bibitem{tanha2020boosting}
J.~Tanha, Y.~Abdi, N.~Samadi, N.~Razzaghi, M.~Asadpour, Boosting methods for
  multi-class imbalanced data classification: an experimental review, Journal
  of Big Data 7~(1) (2020) 1--47.
\newblock \href {https://doi.org/10.1186/s40537-020-00349-y}
  {\path{doi:10.1186/s40537-020-00349-y}}.

\bibitem{alghamdi2017predicting}
M.~Alghamdi, M.~Al-Mallah, S.~Keteyian, C.~Brawner, J.~Ehrman, S.~Sakr,
  Predicting diabetes mellitus using smote and ensemble machine learning
  approach: The henry ford exercise testing (fit) project, PloS one 12~(7)
  (2017) 1--15.
\newblock \href {https://doi.org/10.1371/journal.pone.0179805}
  {\path{doi:10.1371/journal.pone.0179805}}.

\bibitem{chawla2002smote}
N.~V. Chawla, K.~W. Bowyer, L.~O. Hall, W.~P. Kegelmeyer, Smote: synthetic
  minority over-sampling technique, Journal of artificial intelligence research
  16 (2002) 321--357.
\newblock \href {https://doi.org/10.1613/jair.953}
  {\path{doi:10.1613/jair.953}}.

\bibitem{branco2017smogn}
P.~Branco, L.~Torgo, R.~P. Ribeiro, Smogn: a pre-processing approach for
  imbalanced regression, in: First international workshop on learning with
  imbalanced domains: Theory and applications, PMLR, 2017, pp. 36--50.

\bibitem{steininger2021density}
M.~Steininger, K.~Kobs, P.~Davidson, A.~Krause, A.~Hotho, Density-based
  weighting for imbalanced regression, Machine Learning 110~(8) (2021)
  2187--2211.
\newblock \href {https://doi.org/10.1007/s10994-021-06023-5}
  {\path{doi:10.1007/s10994-021-06023-5}}.

\bibitem{ozdemirEEgEmotion2021}
M.~A. Ozdemir, M.~Degirmenci, E.~Izci, A.~Akan, Eeg-based emotion recognition
  with deep convolutional neural networks, Biomedical Engineering /
  Biomedizinische Technik 66~(1) (2021) 43--57.
\newblock \href {https://doi.org/10.1515/bmt-2019-0306}
  {\path{doi:10.1515/bmt-2019-0306}}.

\bibitem{kaur2019diagnosis}
P.~Kaur, M.~Sharma, Diagnosis of human psychological disorders using supervised
  learning and nature-inspired computing techniques: a meta-analysis, Journal
  of medical systems 43~(7) (2019) 1--30.
\newblock \href {https://doi.org/10.1007/s10916-019-1341-2}
  {\path{doi:10.1007/s10916-019-1341-2}}.

\bibitem{christodoulou2019systematic}
E.~Christodoulou, J.~Ma, G.~S. Collins, E.~W. Steyerberg, J.~Y. Verbakel,
  B.~Van~Calster, A systematic review shows no performance benefit of machine
  learning over logistic regression for clinical prediction models, Journal of
  clinical epidemiology 110 (2019) 12--22.
\newblock \href {https://doi.org/10.1016/j.jclinepi.2019.02.004}
  {\path{doi:10.1016/j.jclinepi.2019.02.004}}.

\bibitem{xu2018bayesian}
S.~Xu, Bayesian na{\"\i}ve bayes classifiers to text classification, Journal of
  Information Science 44~(1) (2018) 48--59.
\newblock \href {https://doi.org/10.1177/0165551516677946}
  {\path{doi:10.1177/0165551516677946}}.

\bibitem{zhang2007ml}
M.-L. Zhang, Z.-H. Zhou, Ml-knn: A lazy learning approach to multi-label
  learning, Pattern recognition 40~(7) (2007) 2038--2048.
\newblock \href {https://doi.org/10.1016/j.patcog.2006.12.019}
  {\path{doi:10.1016/j.patcog.2006.12.019}}.

\bibitem{charbuty2021classification}
B.~Charbuty, A.~Abdulazeez, Classification based on decision tree algorithm for
  machine learning, Journal of Applied Science and Technology Trends 2~(01)
  (2021) 20--28.
\newblock \href {https://doi.org/10.38094/jastt20165}
  {\path{doi:10.38094/jastt20165}}.

\bibitem{hengl2018random}
T.~Hengl, M.~Nussbaum, M.~N. Wright, G.~B. Heuvelink, B.~Gr{\"a}ler, Random
  forest as a generic framework for predictive modeling of spatial and
  spatio-temporal variables, PeerJ 6 (2018) e5518.
\newblock \href {https://doi.org/10.7717/peerj.5518}
  {\path{doi:10.7717/peerj.5518}}.

\bibitem{sorensen2018ensemble}
L.~S{\o}rensen, M.~Nielsen, A.~D.~N. Initiative, et~al., Ensemble support
  vector machine classification of dementia using structural mri and
  mini-mental state examination, Journal of neuroscience methods 302 (2018)
  66--74.
\newblock \href {https://doi.org/10.1016/j.jneumeth.2018.01.003}
  {\path{doi:10.1016/j.jneumeth.2018.01.003}}.

\bibitem{rahman2020performance}
S.~Rahman, M.~Irfan, M.~Raza, K.~Moyeezullah~Ghori, S.~Yaqoob, M.~Awais,
  Performance analysis of boosting classifiers in recognizing activities of
  daily living, International journal of environmental research and public
  health 17~(3) (2020) 1082.
\newblock \href {https://doi.org/10.3390/ijerph17031082}
  {\path{doi:10.3390/ijerph17031082}}.

\bibitem{shipe2019developing}
M.~E. Shipe, S.~A. Deppen, F.~Farjah, E.~L. Grogan, Developing prediction
  models for clinical use using logistic regression: an overview, Journal of
  thoracic disease 11~(Suppl 4) (2019) S574--S584.
\newblock \href {https://doi.org/10.21037/jtd.2019.01.25}
  {\path{doi:10.21037/jtd.2019.01.25}}.

\bibitem{fahrmeir2021regression}
L.~Fahrmeir, T.~Kneib, S.~Lang, B.~D. Marx, Regression models, in: Regression,
  Springer, 2021, pp. 23--84.
\newblock \href {https://doi.org/10.1007/978-3-662-63882-8_2}
  {\path{doi:10.1007/978-3-662-63882-8_2}}.

\bibitem{schratz2019hyperparameter}
P.~Schratz, J.~Muenchow, E.~Iturritxa, J.~Richter, A.~Brenning, Hyperparameter
  tuning and performance assessment of statistical and machine-learning
  algorithms using spatial data, Ecological Modelling 406 (2019) 109--120.
\newblock \href {https://doi.org/10.1016/j.ecolmodel.2019.06.002}
  {\path{doi:10.1016/j.ecolmodel.2019.06.002}}.

\bibitem{degirmenci2021ECG}
M.~Degirmenci, M.~Ozdemir, E.~Izci, A.~Akan, Arrhythmic heartbeat
  classification using 2d convolutional neural networks, IRBM (2021).
\newblock \href {https://doi.org/10.1016/j.irbm.2021.04.002}
  {\path{doi:10.1016/j.irbm.2021.04.002}}.

\bibitem{andonie2019hyperparameter}
R.~Andonie, Hyperparameter optimization in learning systems, Journal of
  Membrane Computing 1~(4) (2019) 279--291.
\newblock \href {https://doi.org/10.1007/s41965-019-00023-0}
  {\path{doi:10.1007/s41965-019-00023-0}}.

\bibitem{Olson2016EvoBio}
R.~S. Olson, R.~J. Urbanowicz, P.~C. Andrews, N.~A. Lavender, L.~C. Kidd, J.~H.
  Moore, Applications of evolutionary computation: 19th european conference,
  evoapplications 2016, porto, portugal, march 30 -- april 1, 2016,
  proceedings, part i, Springer International Publishing, 2016, pp. 123--137.
\newblock \href {https://doi.org/10.1007/978-3-319-31204-0_9}
  {\path{doi:10.1007/978-3-319-31204-0_9}}.

\bibitem{ippolito2022hyperparameter}
P.~P. Ippolito, Hyperparameter tuning, in: Applied Data Science in Tourism:
  Interdisciplinary Approaches, Methodologies, and Applications, Springer,
  2022, pp. 231--251.
\newblock \href {https://doi.org/10.1007/978-3-030-88389-8_12}
  {\path{doi:10.1007/978-3-030-88389-8_12}}.

\bibitem{ozdemir2021classification}
M.~A. Ozdemir, G.~D. Ozdemir, O.~Guren, Classification of covid-19
  electrocardiograms by using hexaxial feature mapping and deep learning, BMC
  Medical Informatics and Decision Making 21~(1) (2021) 1--20.
\newblock \href {https://doi.org/10.1186/s12911-021-01521-x}
  {\path{doi:10.1186/s12911-021-01521-x}}.

\bibitem{mcguinness2010comparative}
K.~McGuinness, N.~E. O’connor, A comparative evaluation of interactive
  segmentation algorithms, Pattern Recognition 43~(2) (2010) 434--444.
\newblock \href {https://doi.org/10.1016/j.patcog.2009.03.008}
  {\path{doi:10.1016/j.patcog.2009.03.008}}.

\bibitem{tohka2021evaluation}
J.~Tohka, M.~Van~Gils, Evaluation of machine learning algorithms for health and
  wellness applications: A tutorial, Computers in Biology and Medicine 132
  (2021) 104324.
\newblock \href {https://doi.org/10.1016/j.compbiomed.2021.104324}
  {\path{doi:10.1016/j.compbiomed.2021.104324}}.

\bibitem{altmann2010permutation}
A.~Altmann, L.~Tolo{\c{s}}i, O.~Sander, T.~Lengauer, Permutation importance: a
  corrected feature importance measure, Bioinformatics 26~(10) (2010)
  1340--1347.
\newblock \href {https://doi.org/10.1093/bioinformatics/btq134}
  {\path{doi:10.1093/bioinformatics/btq134}}.

\bibitem{ozdemirHand2022}
M.~A. Ozdemir, D.~H. Kisa, O.~Guren, A.~Akan, Hand gesture classification using
  time–frequency images and transfer learning based on cnn, Biomedical Signal
  Processing and Control 77 (2022) 103787.
\newblock \href {https://doi.org/10.1016/j.bspc.2022.103787}
  {\path{doi:10.1016/j.bspc.2022.103787}}.

\bibitem{steiger1980tests}
J.~H. Steiger, Tests for comparing elements of a correlation matrix.,
  Psychological bulletin 87~(2) (1980) 245.
\newblock \href {https://doi.org/10.1037/0033-2909.87.2.245}
  {\path{doi:10.1037/0033-2909.87.2.245}}.

\bibitem{haghighi2018pycm}
S.~Haghighi, M.~Jasemi, S.~Hessabi, A.~Zolanvari, Pycm: Multiclass confusion
  matrix library in python, Journal of Open Source Software 3~(25) (2018) 729.
\newblock \href {https://doi.org/10.21105/joss.00729}
  {\path{doi:10.21105/joss.00729}}.

\bibitem{Monetta_2011}
T.~Monetta, A.~Scala, C.~Malmo, F.~Bellucci, Antibacterial activity of cold
  plasma-treated titanium alloy, Plasma Medicine 1~(3-4) (2011) 205--214.
\newblock \href {https://doi.org/10.1615/PlasmaMed.v1.i3-4.30}
  {\path{doi:10.1615/PlasmaMed.v1.i3-4.30}}.

\bibitem{bonzanini2021perspectives}
A.~D. Bonzanini, K.~Shao, A.~Stancampiano, D.~B. Graves, A.~Mesbah,
  Perspectives on machine learning-assisted plasma medicine: Toward automated
  plasma treatment, IEEE Transactions on Radiation and Plasma Medical Sciences
  6~(1) (2022) 16--32.
\newblock \href {https://doi.org/10.1109/TRPMS.2021.3055727}
  {\path{doi:10.1109/TRPMS.2021.3055727}}.

\bibitem{Sakai_2022}
O.~Sakai, S.~Kawaguchi, T.~Murakami, Complexity visualization, dataset
  acquisition, and machine-learning perspectives for low-temperature plasma: a
  review, Japanese Journal of Applied Physics 61~(7) (2022) 070101.
\newblock \href {https://doi.org/10.35848/1347-4065/ac76fa}
  {\path{doi:10.35848/1347-4065/ac76fa}}.

\bibitem{fleuren2020machine}
L.~M. Fleuren, T.~L. Klausch, C.~L. Zwager, L.~J. Schoonmade, T.~Guo, L.~F.
  Roggeveen, E.~L. Swart, A.~R. Girbes, P.~Thoral, A.~Ercole, et~al., Machine
  learning for the prediction of sepsis: a systematic review and meta-analysis
  of diagnostic test accuracy, Intensive care medicine 46~(3) (2020) 383--400.
\newblock \href {https://doi.org/10.1007/s00134-019-05872-y}
  {\path{doi:10.1007/s00134-019-05872-y}}.

\bibitem{garg2021role}
A.~Garg, V.~Mago, Role of machine learning in medical research: A survey,
  Computer Science Review 40 (2021) 100370.
\newblock \href {https://doi.org/10.1016/j.cosrev.2021.100370}
  {\path{doi:10.1016/j.cosrev.2021.100370}}.

\bibitem{li2021effective}
C.~Li, C.~Liao, X.~Meng, H.~Chen, W.~Chen, B.~Wei, P.~Zhu, Effective analysis
  of inpatient satisfaction: the random forest algorithm, Patient preference
  and adherence 15 (2021) 691--703.

\bibitem{xiang2022review}
Q.~Xiang, L.~Fan, Y.~Li, S.~Dong, K.~Li, Y.~Bai, A review on recent advances in
  plasma-activated water for food safety: Current applications and future
  trends, Critical Reviews in Food Science and Nutrition 62~(8) (2022)
  2250--2268.
\newblock \href {https://doi.org/10.1080/10408398.2020.1852173}
  {\path{doi:10.1080/10408398.2020.1852173}}.

\bibitem{verlackt2018transport}
C.~Verlackt, W.~Van~Boxem, A.~Bogaerts, Transport and accumulation of plasma
  generated species in aqueous solution, Physical Chemistry Chemical Physics
  20~(10) (2018) 6845--6859.
\newblock \href {https://doi.org/10.1039/C7CP07593F}
  {\path{doi:10.1039/C7CP07593F}}.

\bibitem{lee2019antibacterial}
M.-J. Lee, J.-S. Kwon, H.~B. Jiang, E.~H. Choi, G.~Park, K.-M. Kim, The
  antibacterial effect of non-thermal atmospheric pressure plasma treatment of
  titanium surfaces according to the bacterial wall structure, Scientific
  reports 9~(1) (2019) 1--13.
\newblock \href {https://doi.org/10.1038/s41598-019-39414-9}
  {\path{doi:10.1038/s41598-019-39414-9}}.

\end{thebibliography}

\newpage

\begin{landscape}

\begin{table}
\begin{center}
\begin{minipage}{\textwidth}
\caption{Input and output variables, their categories, types, and the total numbers of unique values (A detailed expression of unique values of nominal variables is presented in Table S2 in Supplementary Data).}
\label{table:predictors}
\begin{tabular}{llcc }
\toprule
\textbf{Category} & \textbf{Variables} & \textbf{Type} & \textbf{Unique Value} \\
\toprule 
\multirow{4}{10em}{\textbf{CAP Specifications}} & {Plasma Treatment Type} & Nominal & 7 \\
{} & {Gas Type} & Nominal & 8 \\
{} & {Discharge Gap} & Numeric & 15 \\
{} & {Plasma Treatment Time} & Numeric & 21 \\
\midrule
\multirow{2}{10em}{\textbf{PAL Characteristics}} & {Plasma Activated Liquid} & Nominal & 15 \\
{} & {Treatment Volume} & Numeric & 14 \\
\midrule
\multirow{2}{10em}{\textbf{In vitro Characteristics}} & {Microbial Strain} & Nominal & 22 \\
{} & {Initial Microbial Load} & Numeric & 32 \\
\midrule
\multirow{4}{10em}{\textbf{CAP Treatment Characteristics}} & {PAL/mo Suspension Volume Ratio} & Numeric & 12 \\
{} & {Contact Time} & Numeric & 26 \\
{} & {Incubation Temperature} & Numeric & 8 \\
{} & {Post Storage Time} & Numeric & 27 \\
\midrule
\multirow{2}{10em}{\textbf{Output}} & \multirow{2}{10em}{Microbial Inactivation} & Numeric & 126 \\
{} & {} & Categorical & 4 \\
\bottomrule
\end{tabular}
\end{minipage}
\end{center}
\end{table}

\begin{table}
\begin{center}
\begin{minipage}{\textwidth}
\caption{Categorization conditions for MI (Microbial Inactivation: Outcome) and counts of class labels before applying SMOTE (\textit{n} indicates the initial microbial load).}
\label{table:MI}
\begin{tabular}{ccc}
\toprule
\textbf{Condition} & \textbf{Output Label} & \textbf{Counts} \\
\toprule 
{\textbf{${MI\le{0.1n}}$}} & {None, N} & {288} \\
{\textbf{${0.1n<MI<0.5n}$}} & {Weak, W} & {169} \\
{\textbf{${0.5n\le{MI}<0.9n}$}} & {Strong, S} & {94} \\
{\textbf{${MI\ge{0.9n}}$}} & {Complete, C} & {211} \\
\bottomrule
\end{tabular}
\end{minipage}
\end{center}
\end{table}

\begin{table}
\begin{center}
\begin{minipage}{\textwidth}
\caption{Descriptive statistics of numeric variables.}
\label{table:statistics}
\begin{tabular}{lccccccc }
\toprule
\textbf{Numeric Variable} & \textbf{Unit} & \textbf{Count} & \textbf{Mean} & \textbf{Std.} & \textbf{Min.} & \textbf{Max.}  \\
\toprule 
{\textbf{Discharge Gap}} & mm & 762 & 13.71 & 17.23 & 0 & 81 \\
{\textbf{Plasma Treatment Time}} & sec & 762 & 1165 & 2771.48 & 0 & 14400 \\
{\textbf{Treatment Volume}} & mL & 762 & 45.52 & 116.56 & 0.25 & 500 \\
{\textbf{Initial Microbial Load}} & log & 762 & 6.62 & 0.91 & 2 & 9 \\
{\textbf{PAL/mo Suspension Volume Ratio}} & fold & 762 & 109.29 & 246.75 & 1 & 1000 \\
{\textbf{Contact Time}} & min & 762 & 39.05 & 131.02 & 0 & 1440 \\
{\textbf{Incubation Temperature}} & \textdegree C & 762 & 23.5 & 12.4 & -80 & 50 \\
{\textbf{Post Storage Time}} & hour & 762 & 64.47 & 345.56 & 0 & 6120 \\
{\textbf{Microbial Inactivation}} & log & 762 & 2.83 & 2.77 & 0 & 9 \\
\bottomrule
\end{tabular}
\end{minipage}
\end{center}
\end{table}

\begin{table}
\begin{center}
\begin{minipage}{\textwidth}
\caption{Average performance metrics of 3-repeated stratified-10-fold cross-validation strategy for different classifiers (ACC, F1, REC, and PRE values are given as percentages (\%) and the ET unit is s.).  }
\label{table:classifiermetrics}
\begin{tabular}{l|cc|ccccccccc}
\toprule
\multirow{2}{5em}{\textbf{Classifier}} & \multicolumn{2}{c|}{\textbf{Train}} & \multicolumn{8}{c}{\textbf{Validation}} \\
{} & \textbf{ACC} & \textbf{AUC} & \textbf{ACC} & \textbf{F1} & \textbf{REC} & \textbf{PRE} & \textbf{JI} & \textbf{AUC} & \textbf{ET} & \textbf{\textit{p}-value} \\
\hline 
{LR} & 55.40 & 0.82 & 55.53 & 55.40 & 55.50 & 57.17 & 0.39 & 0.77 & 1.28 & $>.999$ \\
{LDA} & 60.20 & 0.83 & 54.70 & 54.43 & 54.60 & 56.43 & 0.38 & 0.78 & 1.62 & $>.999$ \\
{GPC} & 70.07 & 0.90 & 61.20 & 61.03 & 61.20 & 62.13 & 0.45 & 0.83 & 92.17 & $>.999$ \\
{XGBC} & 98.20 & 1.00 & 80.23 & 80.23 & 80.23 & 80.70 & 0.68 & 0.94 & 9.76 & $<.001$ \\
{LGBMC} & 97.00 & 1.00 & 78.63 & 78.60 & 78.57 & 79.33 & 0.65 & 0.94 & 9.76 & .250\\
{KNN} & 77.40 & 0.95 & 65.20 & 64.93 & 65.37 & 66.23 & 0.49 & 0.85 & 3.16 & $>.999$\\
{DTC} & 99.00 & 1.00 & 74.73 & 74.63 & 74.77 & 75.43 & 0.60 & 0.84 & 1.20 & $>.999$\\
{ETC} & 99.00 & 1.00 & 70.37 & 70.30 & 70.30 & 71.00 & 0.55 & 0.81 & 1.07 & $>.999$\\
{GNB} & 40.52 & 0.76 & 38.17 & 31.80 & 37.97 & 45.57 & 0.20 & 0.72 & 1.14 & $>.999$\\
{BNB} & 51.30 & 0.76 & 46.90 & 46.73 & 46.97 & 47.53 & 0.31 & 0.72 & 1.17 & $>.999$\\
{SVM} & 48.03 & 0.77 & 46.20 & 45.53 & 46.23 & 50.47 & 0.30 & 0.74 & 14.77 & $>.999$\\
{BC} & 97.50 & 1.00 & 77.63 & 77.57 & 77.63 & 78.23 & 0.64 & 0.92 & 2.50 & .781 \\
{ABC} & 62.20 & 0.81 & 58.17 & 58.13 & 58.23 & 59.37 & 0.42 & 0.77 & 6.35 & $>.999$\\
{HGBC} & 97.00 & 1.00 & 78.50 & 78.47 & 78.50 & 79.20 & 0.65 & 0.93 & 132.98 & .331 \\
\textbf{{RFC}} & 99.00 & 1.00 & \textbf{80.47} & 80.47 & 80.47 & 81.03 & 0.68 & 0.94 & 10.50 & $<.001$ \\
{GBC} & 88.00 & 0.98 & 73.47 & 73.43 & 73.47 & 74.37 & 0.59 & 0.91 & 22.27 & $>.999$ \\
\bottomrule
\end{tabular}
\end{minipage}
\end{center}
\end{table}

\begin{table}
\begin{center}
\begin{minipage}{\textwidth}
\caption{Average performance metrics of 10-fold cross-validation strategy for different regressors (ET unit is s.).}
\label{table:regressormetrics}
\begin{tabular}{l|ccc|ccccccc}
\toprule
\multirow{2}{5em}{\textbf{Regressor}} & \multicolumn{3}{c|}{\textbf{Train}} & \multicolumn{6}{c}{\textbf{Validation}} \\
{} & \textbf{R\textsuperscript{2}} & \textbf{MAE} & \textbf{RMSE} & \textbf{R\textsuperscript{2}} & \textbf{MAE} & \textbf{MSE} & \textbf{RMSE} & \textbf{ET} & \textbf{\textit{p}-value} \\
\hline 
{LASSO} & 0.08 & 0.88 & 0.96 & 0.05 & 0.89 & 0.95 & 0.97 & 0.18  & $>.999$ \\
{RR} & 0.28 & 0.72 & 0.86 & 0.13 & 0.79 & 0.87 & 0.93 & 0.17  & $>.999$\\
{XGBR} & 0.96 & 0.10 & 0.19 & 0.68 & 0.36 & 0.32 & 0.56 & 1.35 & .009 \\
{LLars} & 0.19 & 0.80 & 0.91 & 0.13 & 0.82 & 0.87 & 0.93 & 0.23 &$>.999$ \\
{KNR} & 0.59 & 0.47 & 0.64 & 0.37 & 0.62 & 0.63 & 0.79 & 0.23 & $>.999$\\
{ABR} & 0.31 & 0.74 & 0.84 & 0.26 & 0.76 & 0.74 & 0.86 & 0.50 &$>.999$ \\
{ETR} & 0.98 & 0.02 & 0.14 & 0.68 & 0.31 & 0.32 & 0.56 & 3.03 & .003 \\
{BR} & 0.93 & 0.14 & 0.26 & 0.69 & 0.34 & 0.31 & 0.55 & 0.49 & $<.001$ \\
{ENR} & 0.00 & 0.92 & 1.01 & 0.00 & 0.92 & 1.02 & 1.01 & 0.18 &$>.999$ \\
{LSVR} & 0.16 & 0.64 & 0.92 & 0.06 & 0.76 & 1.07 & 1.03 & 0.22  &$>.999$ \\
{BaR} & 0.20 & 0.79 & 0.90 & 0.13 & 0.82 & 0.87 & 0.93 & 0.48  &$>.999$ \\
{MLPR} & 0.56 & 0.50 & 0.66 & 0.28 & 0.65 & 0.72 & 0.84 & 17.45  &$>.999$ \\
{\textbf{RFR}} & 0.95 & 0.13 & 0.23 & \textbf{0.72} & 0.33 & 0.28 & 0.53 & 3.05 & $<.001$\\
{GBR} & 0.73 & 0.42 & 0.52 & 0.58 & 0.52 & 0.42 & 0.64 & 0.99 & .952\\
\bottomrule
\end{tabular}
\end{minipage}
\end{center}
\end{table}


\begin{figure}
\begin{center}
\includegraphics[height=0.75\textheight]{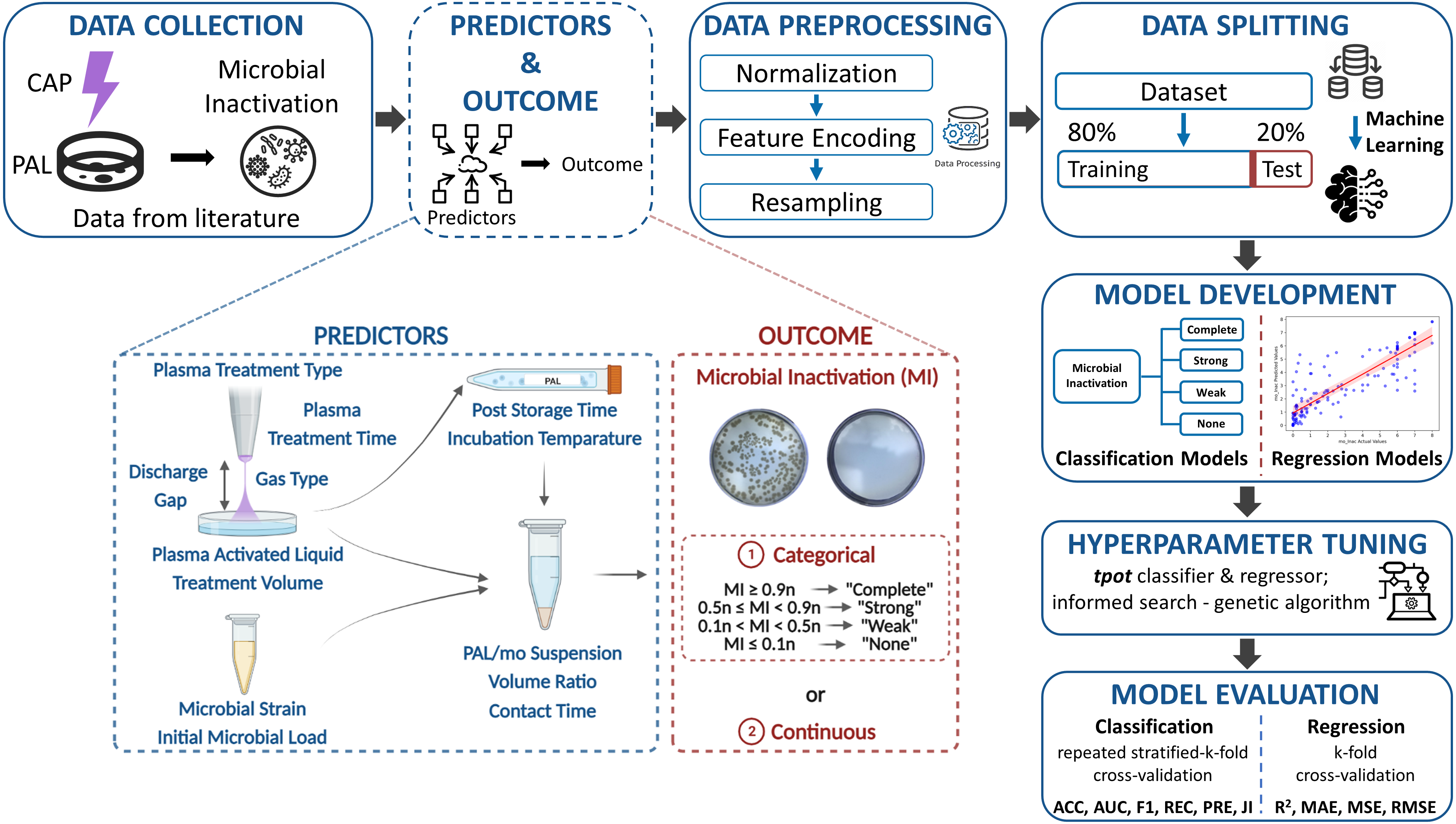}
\caption{The illustration of the proposed study framework.}
      \label{Fig:framework}
\end{center}
\end{figure}            

\end{landscape}

\end{document}